\documentclass[12pt,lot,lof]{puthesis_undergraduate}
\usepackage{amsfonts}
\usepackage{amssymb}
\usepackage{amsmath}
\usepackage{amsthm}
\usepackage{latexsym}
\usepackage{graphicx}
\usepackage{dirtree}
\usepackage{url}
\newcommand{\bes}{\begin{equation*}}

\title{Building Multimodal AI Chatbots}
\submitted{May 2023}  %graduation date
\author{Min Young Lee}
\advisor{Professor Jia Deng} %
\abstract{

This work aims to create a multimodal AI system that chats with humans and shares relevant photos. While earlier works were limited to dialogues about specific objects or scenes within images, recent works have incorporated images into open-domain dialogues. However, their response generators are unimodal, accepting text input but no image input, thus prone to generating responses contradictory to the images shared in the dialogue. Therefore, this work proposes a complete chatbot system using two multimodal deep learning models: an image retriever that understands texts and a response generator that understands images. The image retriever, implemented by ViT and BERT, selects the most relevant image given the dialogue history and a database of images. The response generator, implemented by ViT and GPT-2/DialoGPT, generates an appropriate response given the dialogue history and the most recently retrieved image. The two models are trained and evaluated on PhotoChat, an open-domain dialogue dataset in which a photo is shared in each session. In automatic evaluation, the proposed image retriever outperforms existing baselines VSE++ and SCAN with Recall@1/5/10 of 0.1/0.3/0.4 and MRR of 0.2 when ranking 1,000 images. The proposed response generator also surpasses the baseline Divter with PPL of 16.9, BLEU-1/2 of 0.13/0.03, and Distinct-1/2 of 0.97/0.86, showing a significant improvement in PPL by $-42.8$ and BLEU-1/2 by $+0.07/0.02$. In human evaluation with a Likert scale of 1-5, the complete multimodal chatbot system receives higher image-groundedness of 4.3 and engagingness of 4.3, along with competitive fluency of 4.1, coherence of 3.9, and humanness of 3.1, when compared to other chatbot variants. The source code is available at: \url{https://github.com/minniie/multimodal_chat.git}.

}

\acknowledgements{
I would like to thank Professor Jia Deng in the Department of Computer Science at Princeton University. Completing this thesis after coming back from two years of leave of absence was a rough journey, and Professor Deng gave his full advice and support.

I extend my sincere thanks to Professor Sebastian Seung in the Department of Computer Science and Princeton Neuroscience Institute at Princeton University for kindly accepting to be the second reader of this thesis. His class on neural networks helped me gain a thorough theoretical understanding of the deep learning models I used in this project.

I also express gratitude to my coworkers at Generative Chatbot Team of NAVER Corp: Donghyun Kwak, Sanghwan Bae, Soyoung Kang, and Donghoon Ham. Our team's research on open-domain dialogue and generative AI motivated me to start this topic in the first place, and all of them were extremely supportive while I was temporarily away for finishing up my senior year in Princeton.

Last but not least, this thesis would not have been possible without the unconditional love from my family in Korea: my mom, my dad, my little sister, and our cat May.
}

\begin{document}

\chapter{Introduction}\label{ch:introduction}
\section{Motivation and Goal}

The overarching goal of artificial intelligence is to imitate human intelligence, which includes the abilities to perceive sensory information, communicate with others, create novel ideas, and much more. Most of the early AI systems have tried to implement such abilities by tackling various tasks in each of the visual, acoustic, linguistic, and spatial modalities independently. Recently, many works have started integrating these modalities and proposed systems that process inputs from two or more sources.

%Due to the rapid development of deep learning technologies,

\begin{figure}[!hbt]
\centering
\setlength{\belowcaptionskip}{15pt}
\includegraphics[width=0.9\linewidth]{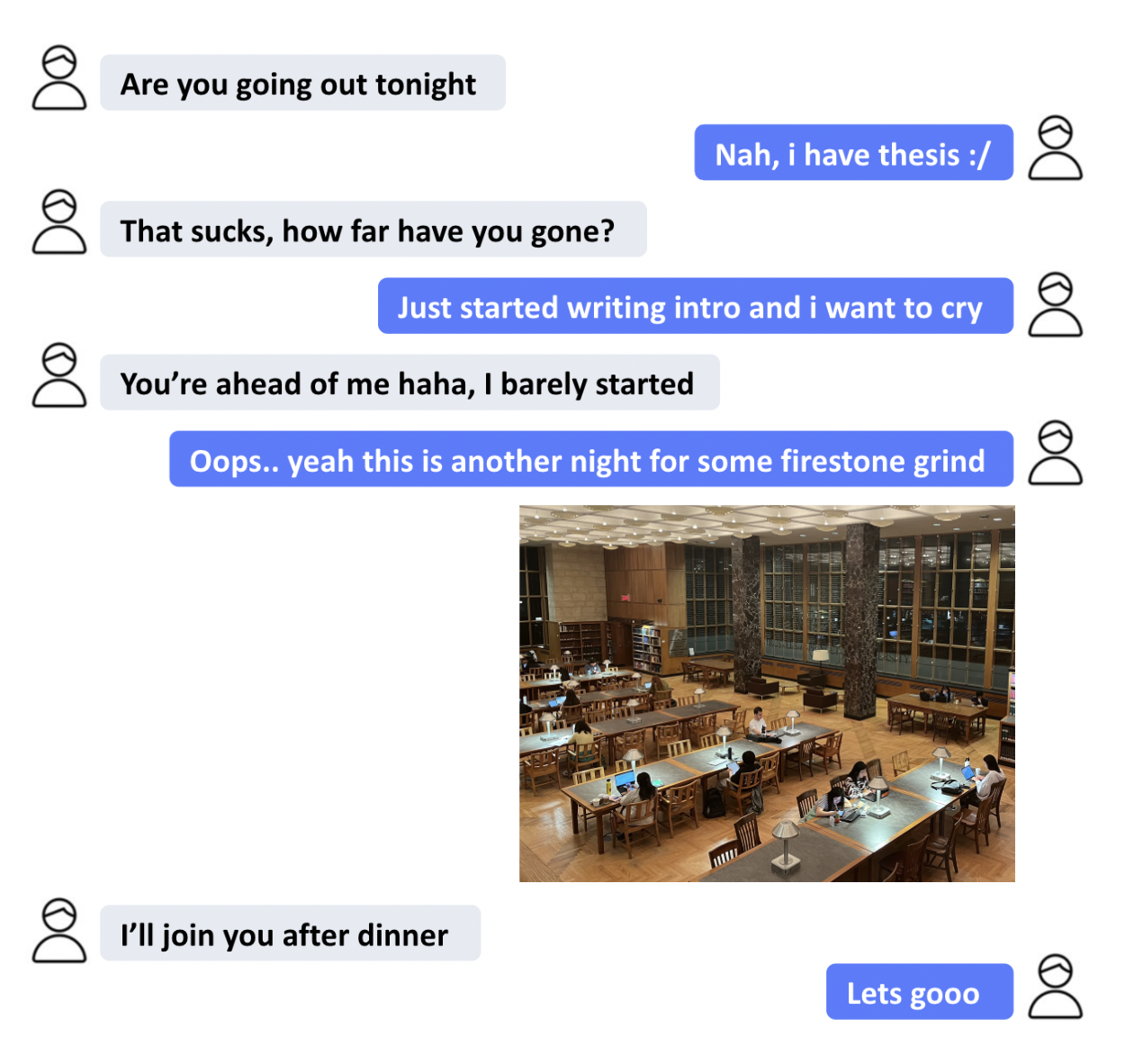}
\caption{A conversation between two speakers in which a photo is shared.}
\label{fig:human_dialogue}
\end{figure}

Given the current state of the research field, the goal of this work is to build an AI system that is able to converse with users and send relevant photos. This is a natural direction in emulating human intelligence, since people often communicate by sending messages and photos through chat applications. An example is illustrated in Figure \ref{fig:human_dialogue}. Two people are complaining about writing senior thesis in Princeton, and one of them sends a picture of the Firestone Library at one point in the conversation. Here, the speaker naturally possesses the abilities to (1) pick out a relevant photo from their gallery and (2) write message that is coherent to the chosen photo. This work aims to model these two abilities and create an interactive chatbot system using deep learning models.

%The goal of this work is to build an open-domain photo-sending chatbot system.
%It is open-domain, in that the user can chat with the system about possibly any topics, and photo-sending, in that the chatbot will send relevant photos during the conversation.
%This task requires image-text multimodal AI modeling, which is a recently burgeoning field in AI, while this specific task was never fully explored.
%This system has many potential applications, including virtual AI friends.
%Figure \ref{fig:human_dialogue} illustrates a natural dialogue between humans, in which a relevant photo is shared by one speaker.
%The recent popularity of ChatGPT is an evidence of how engaging and helpful a conversational AI system can be.

\section{Previous Works}

Several works have aimed to model image and dialogue together. The first line of such work is image-grounded dialogue (Section \ref{Image-grounded Dialogue}), which is the task of generating an answer to an open-ended question about some object or scene in each image. However, this task is not sufficient for the goal of my work, since my chatbot is intended to carry out a conversation about any topic even in the absence of an image.

The second line of work is image-augmented dialogue (Section \ref{Image-augmented Dialogue}). The task is to generate a proper response to any dialogue context in which an image may or may not be shared, which aligns with the objective of my work. However, the methods proposed in these works do not fully meet my criteria. One work \cite{zang-etal-2021-photochat} releases a dataset on this task and implements image-sharing classifier and image retriever, but it does not build any models that generate responses, which is required for any chatbot that can converse with humans. A later work \cite{sun-etal-2022-multimodal} proposes a complete chatbot system with response generator and image translator. However, this response generator is unimodal and only takes in dialogue history as model input, which means that it does not understand any previously retrieved images and can generate responses contradictory to those images.

%However, the current approaches in image-augmented dialogue and chatbot are lacking in a few aspects.
%One approach \cite{sun-etal-2022-multimodal} implements an image-augmented chatbot by using a response generator model and a image generator model.
%A brief sketch of this chatbot pipeline is as follows: given the dialogue context of previous conversation with user and bot, at each timestep, the response generator generates a dialogue response and the image generator generates an appropriate image.
%However, this response generator is a unimodal model, meaning that it does not take image as the input.
%This indicates that once an image is shared within a dialogue, the response generator will ignore any image features from that image and will continue the dialogue using only the textual history.

\section{Approach}

The proposed chatbot system consists of an image model and a dialogue model. The fundamental idea behind my approach is that the image model and the dialogue model should both be \textit{multimodal}: the image model should understand dialogue and the dialogue model should understand image.

The image model is implemented as an image retriever. This image retriever takes in both the dialogue history and a database of images as input, and returns the most appropriate image as output if its score is greater than a predefined threshold. The dialogue model is implemented as a response generator. This response generator takes in both the dialogue history and the most recently retrieved image as input, and returns a text response as output. Whenever a user sends message to the chatbot, these two models are serially called. The generated response and optionally the retrieved image are sent to the user.

%The main intuition behind my approach is that the dialogue model and the image model should be \textit{multi-modal to each other} -- that is, the dialogue model should understand image, and the image model should understand dialogue.
%My approach consists of two models: image retriever (Section \ref{Approach:Image Retriever}) and response generator (Section \ref{Approach:Response Generator}).
%This model takes in the dialogue history as the text feature and a predefined pool of images as the image feature, and outputs the most appropriate image to the context.
%This model takes in the dialogue history as the text feature and the previously shared image (if applicable) as the image feature, and outputs a response to the user.
%The two models create the proposed photo-sending open-domain chatbot pipeline.

\section{Implementation}

The multimodal image retriever consists of an image encoder and a text encoder (Section \ref{Implementation: Image Retriever}). The image encoder encodes each image into an image representation, and the text encoder encodes the dialogue history into a text representation. The two representations are mapped to a joint image-text representation space, and their cosine similarity is computed. Out of all images in the database, the image with maximum cosine similarity is retrieved. ViT \cite{DBLP:journals/corr/abs-2010-11929} is used as image encoder and BERT \cite{devlin-etal-2019-bert} is used as text encoder. A contrastive loss is used for learning.

The multimodal response generator consists of an image encoder and a text decoder (Section \ref{Implementation: Response Generator}). The image encoder encodes the retrieved image into an image representation. The text decoder encodes the dialogue history and previously generated tokens into a text representation, combines this text representation with the image representation into a joint representation, and maps it onto a vocabulary space. Then, a token is sampled from this probability over vocabulary space, and this token is appended to dialogue history for generating subsequent tokens. ViT \cite{DBLP:journals/corr/abs-2010-11929} is used as image encoder and GPT-2 \cite{radford2019language} or DialoGPT \cite{zhang-etal-2020-dialogpt} is used as text decoder. A cross entropy loss is used for learning.

\section{Results}

The proposed system is trained and evaluated using PhotoChat \cite{zang-etal-2021-photochat}, an open-domain dialogue dataset in which a photo is shared in each session (Section \ref{Evaluation: Dataset}). The system is evaluated using both automatic evaluation and human evaluation methods.

First, the image retrievers are automatically evaluated using Recall@1/5/10 and MRR (Section \ref{Automatic Evaluation: Image Retriever}). All four proposed image retriever variants implemented with ViT and BERT achieve approximately 0.1/0.3/0.4 in Recall@1/5/10 and 0.2 in MRR when ranking 1,000 candidates. Also, the four models surpass two baseline models VSE++ \cite{faghri2018vse} and SCAN \cite{lee2018stacked} in Recall@5/10 with maximum difference of $+0.06/0.09$ and $+0.04/0.06$, respectively, showing the effectiveness of transformer-based dual encoders.

Second, the response generators are automatically evaluated using PPL, BLEU-1/2 \cite{papineni-etal-2002-bleu}, and Distinct-1/2 \cite{li-etal-2016-diversity} (Section \ref{Automatic Evaluation: Response Generator}). All six proposed response generator variants implemented with ViT and GPT-2 or DialoGPT significantly outperform an existing baseline Divter \cite{sun-etal-2022-multimodal} by $-42.8$ in PPL at maximum and $+0.07/0.02$ in BLEU-1/2 on average. Also, among the proposed models, the multimodal variants achieve PPL of 16.9, lower than that of the unimodal variants roughly by $-10.8$, which indicates that giving both image and text inputs to response generators is helpful in predicting image-augmented responses. The multimodal variants also exhibit BLEU-1/2 of 0.13/0.03 and Distinct-1/2 of 0.97/0.85, a result comparable to the unimodal variants.

Finally, the complete chatbot system is evaluated by human crowdworkers through turn evaluation and session evaluation (Section \ref{Human Evaluation: Chatbot}). While conversing with the deployed chatbot, each crowdworker rates each response or entire dialogue using a Likert scale of 1-5 to assess fluency, coherence, image-groundedness, engagingness, and humanness. In turn evaluation, the proposed chatbot with multimodal image retriever and response generator achieves fluency of 4.1 and coherence of 3.9, similarly to other chatbot variants either with unimodal response generator or without any image retriever. This chatbot also outperforms the unimodal variant in image-groundedness by $+0.3$ with an absolute score of 4.3, which aligns with the hypothesis that response generators that additionally understand images are more capable of generating responses consistent with the shared image. Furthermore, in session evaluation, the proposed multimodal chatbot achieves the highest engagingness of 4.3, with margins of $+0.1$ and $+0.6$ compared to the chatbots with unimodal response generator and without image retriever, respectively. This signifies that chatbots that understand and send images properly can improve the user experience of conversational AI systems. All three chatbot variants score 3.1 in humanness, which suggests a room for improvement in making chatbots more like humans. 

\chapter{Previous Works}
\section{Image-grounded Dialogue} \label{Image-grounded Dialogue}

One of the initial tasks in joint modeling of image and dialogue is image-grounded dialogue. This task is to generate proper answers to questions about the image. One work \cite{mostafazadeh-etal-2017-image} proposes a method to generate pairs of open-ended question and answer about an image and uses this method to build IGC, a dataset of single-turn dialogues grounded on images. For example, given an image of a flat tire, the generated question is ``Do you think this happened on the highway?" and the generated answer is ``Probably not, because I haven’t driven anywhere except around town recently." Another work \cite{shuster-etal-2020-image} expands this task to multi-turn and releases Image-Chat, a dataset in which two speakers talk about some object in the image over multiple turns. Each speaker in Image-Chat has additional style label, such as peaceful, absentminded, or miserable, and adheres to this style when speaking about the object.

The target task in these two works \cite{mostafazadeh-etal-2017-image, shuster-etal-2020-image}, however, is not precisely the goal of this project, because it concerns only dialogues \textit{grounded on} images.
Although both IGC and Image-Chat contain various images, their conversations primarily involve talking about some object, person, or scene in each image and rarely extend to non-image topics.
My goal, on the other hand, is to cover dialogues open to any possible topic, independent of what types of images are available.
Unlike the works \cite{mostafazadeh-etal-2017-image, shuster-etal-2020-image} that deal with image-\textit{grounded} dialogues, my work concerns image-\textit{augmented} dialogues, in which images rather serve as an additional component to improve the liveliness of open-domain conversations. 

\section{Image-augmented Dialogue} \label{Image-augmented Dialogue}

The task of utilizing images in a completely open conversational setting has recently been targeted in a few previous studies. One work \cite{zang-etal-2021-photochat} releases PhotoChat, the first dialogue dataset in which a photo is shared at some turn in each session. This work also introduces two models needed in an image-augmented chatbot system: image-sharing intent classifier and image retriever. The image-sharing intent classifier classifies whether the current dialogue history is suitable for sharing some image, and if this classifier predicts positive, then the image retriever retrieves the image most appropriate to the dialogue history from a database of images.

Although these two models execute core functions of a photo-sending chatbot system, this work \cite{zang-etal-2021-photochat} does not implement any response generator models and only evaluates the performance of the two models independently using the test set of PhotoChat. Thus, the actual chatbot that is able to carry out the conversation is missing.

A later work \cite{sun-etal-2022-multimodal} proposes a pipeline of response generator and image translator that can send both responses and photos to users. To the best of my knowledge, this is the only existing work that implements a full chatbot system with photo sending abilities. In this work, the response generator takes in the dialogue history and generates a response and optionally a description of an image that would be appropriate to the context. If the image description is generated, it is given as input to the image translator to generate the corresponding image. The output of the system at each conversation turn is thus the response and optionally the image.

\begin{figure}[!hbt]
\centering
\vspace*{15pt}
\includegraphics[width=1.0\linewidth]{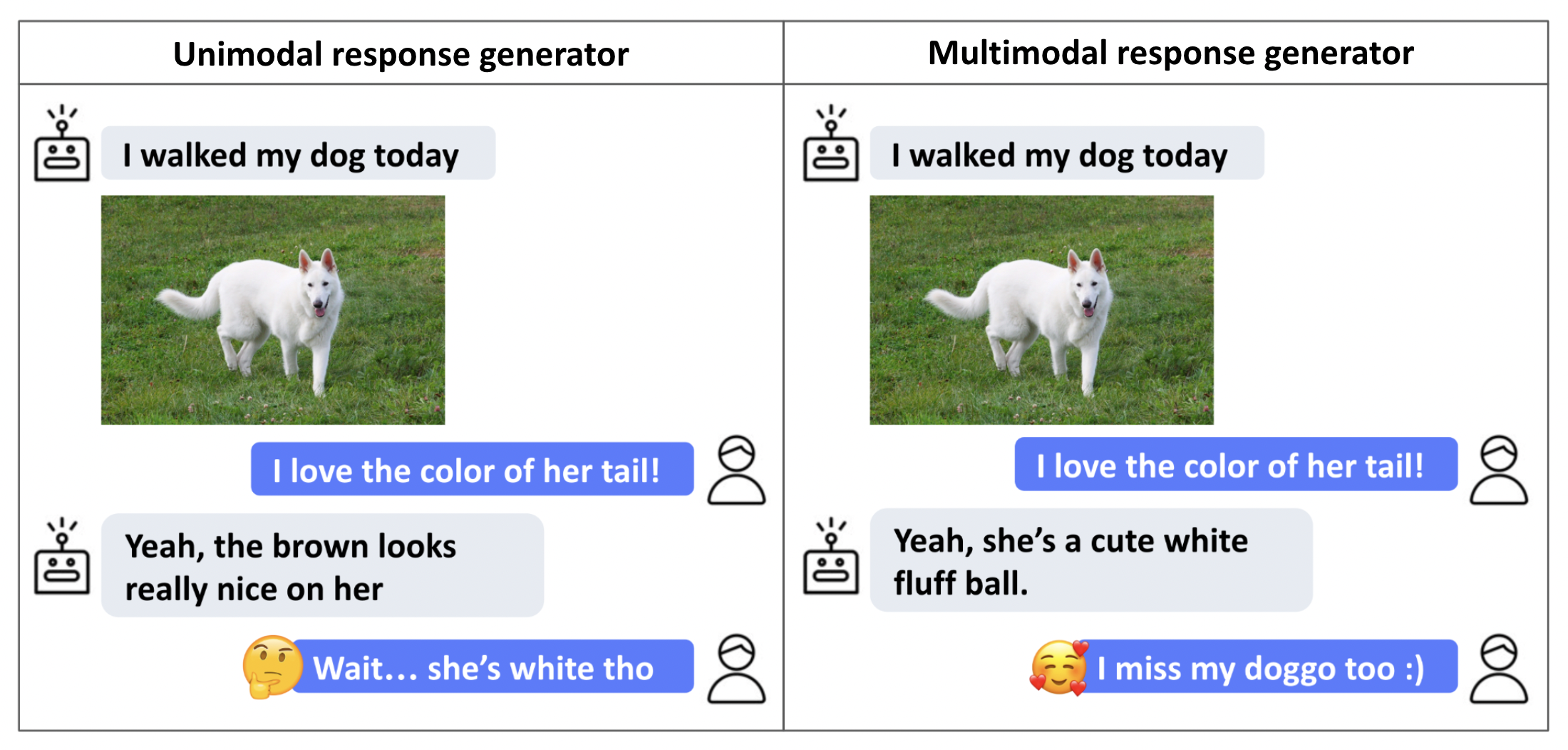}
\caption{Comparison between unimodal generator and multimodal generator. In the left dialogue, the bot fails to understand the previously sent image and says something that is inconsistent with the image. In the right dialogue, the bot understands the image and carries out the conversation naturally.}
\label{fig:image_augmented_chatbot}
\end{figure}

Despite its success in incorporating images into a dialogue system, this work \cite{sun-etal-2022-multimodal} is still limited in that its response generator is a unimodal model. This means that the model takes in only the dialogue history as input, thus is unable to understand images that have been shared previously in the dialogue. An example is shown in Figure \ref{fig:image_augmented_chatbot}. Here, the bot first shares an image of their dog, and the user responds by mentioning the color of the dog's tail. If the response generator of the bot only understands dialogue history, the scenario on the left becomes possible: the bot may hallucinate and respond with something like ``Yeah, the brown looks really nice on her," while the actual color of the dog's tail in the image is white. Thus, in order to avoid such egregious hallucination, the response generator must be a multimodal model, being able to understand both the shared image and the dialogue history and respond coherently like the scenario on the right of Figure \ref{fig:image_augmented_chatbot}. 

%Second, the translator suffers from high latency. A chatbot user expects at most 1 second for a response, but N is too much for good user experience.

\chapter{Approach}
\section{Proposed System}

\begin{figure}[!hbt]
\centering
\setlength{\belowcaptionskip}{15pt}
\includegraphics[width=1.0\linewidth]{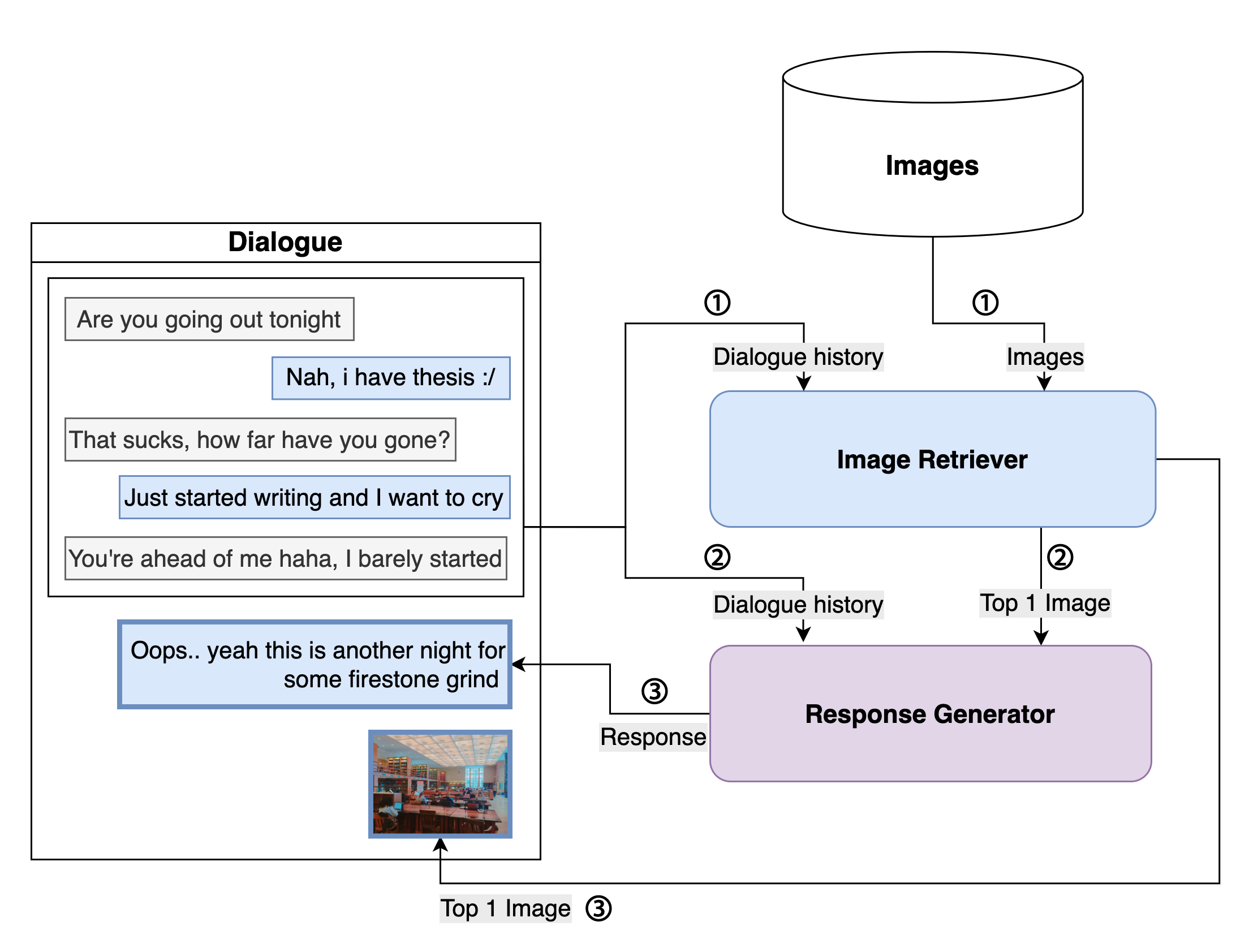}
\caption{The proposed system. Each of the three steps is numbered. For simplicity, it only illustrates the case in which a top 1 image is retrieved by the image retriever.}
\label{fig:system}
\end{figure}

The proposed chatbot system is illustrated in Figure \ref{fig:system}. The system follows three main steps in order to carry out successful conversations with users.

First, the dialogue history and a database of images are fed to the image retriever. The dialogue history is a set of all previous messages between the user and the chatbot, including the most recently sent message by the user. The database of images is a set of thousands of images stored in a remote location and is accessible by the chatbot in real-time. Given these two inputs, the image retriever ranks each image in the database by its similarity score with the dialogue history. If the score of the most similar image is above some predefined threshold, the image retriever returns that image as the top 1 image. If the score is below the threshold, the image retriever returns nothing. %a dummy image of zero pixels.

Second, the same dialogue history and the retrieved image are fed to the response generator. If no image is retrieved from the first step, the most recently shared image in the dialogue is given instead. If no image is retrieved and no images have been shared, a dummy image of all zeros is given. At each timestep, the response generator samples a token from the vocabulary, conditioned on the dialogue history, the given image, and previously generated tokens. It continues sampling tokens autoregressively until the end-of-sequence token is generated. The final response is a concatenation of all sampled tokens.

Third, the response returned by the response generator is sent to the user and displayed on the chatbot interface. If top 1 image is retrieved from the first step, it is also displayed after the response. For subsequent conversations, the response is appended to the dialogue history and the image is stored in a queue of previously shared images. The system remains idle until the user sends the next message, at which the three steps are iterated again.

\chapter{Implementation}
%The system consists of two models: image retriever (Section \ref{Implementation: Image Retriever}) and response generator (Section \ref{Implementation: Response Generator}).

\section{Image Retriever} \label{Implementation: Image Retriever}

This section describes the implementation details of the image retriever, including model architecture, input and output, training and inference, and computational resource.

\subsection{Model Architecture} \label{Implementation: Image retriever: Model architecture}

The base architecture of image retriever is a \verb|VisionTextDualEncoder| from Huggingface \cite{wolf-etal-2020-transformers}, an open-source platform for varieties of recent model architectures. \verb|VisionTextDualEncoder| is a general class for vision-text dual encoder architectures, consisting of an image encoder and a text encoder. This architecture is chosen because any previously released image encoders and text encoders can be plugged into this class, even when each encoder alone may not support multimodal learning. Thus, it is easy to leverage the performance of the state-of-the-art architectures in each independent modality without being restricted to publicly released multimodal architectures.

\verb|VisionTextDualEncoder| consists of two main architectural components: an image encoder and a text encoder. For the task of image retrieval, the image encoder takes in each image in the database and encodes it into an image representation. In parallel, the text encoder takes in the dialogue history up to the current turn and encodes it into a text representation. Then, a fully connected layer on top of each encoder maps each representation to a joint image-text representation space. As the two resulting representations have the same dimension, a cosine similarity can be computed to express how similar each image and the dialogue history are.
%After computing similarity with all images in the database at inference, the image with maximum similarity is selected. If its score is above a predefined threshold, the image is retrieved and shared to the user. If it is below the threshold, no image is retrieved. The threshold is empirically set as 0.2 from preliminary experiments on the validation set.

The family of models used as image encoders in image retrievers is Vision Transformer (ViT) \cite{DBLP:journals/corr/abs-2010-11929}. Transformers \cite{https://doi.org/10.48550/arxiv.1706.03762} are a group of deep learning models that uses attention mechanism to attend over relevant parts of the input to generate the output. Although transformers gained popularity initially in text models, ViT adopted transformer encoder as its backbone architecture for tasks in vision domain and used local patches of images as the input sequence. It is known to achieve significant performance in various image recognition tasks. As input to ViT, a special token named \verb|[class]| is prepended to the sequence of image patches, and the representation of this \verb|[class]| token returned from its final transformer encoder block is interpreted as the representation of the entire image. 

Two variants of ViT are used in the experiment: ViT-base and ViT-large. The two models follow the basic ViT architecture. The main difference is that ViT-large has larger number of trainable parameters than ViT-base, resulting from larger number of transformer encoder blocks, size of hidden dimension, and size of input image. The exact numbers of the parameters are in Appendix \ref{Appendix: Model architecture}. The pretrained checkpoints of ViT are downloaded from Huggingface \cite{wolf-etal-2020-transformers}. The names of the checkpoints for ViT-base and ViT-large are \verb|google/vit-base-patch16-224| and \verb|google/vit-large-patch32-384|, respectively.

On the other hand, the family of models used as text encoders in image retrievers is BERT \cite{devlin-etal-2019-bert}. Similar to ViT \cite{DBLP:journals/corr/abs-2010-11929}, BERT is also based on a transformer encoder. To create the representation of each token, BERT attends over all other tokens from both left and right and computes a weighted sum of the other token representations, in which each weight depends on each attention score. For the purpose of obtaining the representation of the entire dialogue history, a special token named \verb|[CLS]| is prepended to the sequence of input tokens, similarly to ViT. This \verb|[CLS]| representation is interpreted as the representation of the dialogue history.

Two variants of BERT are used in the experiment: BERT-base and BERT-large. The two models follow the basic BERT architecture. The main difference is that BERT-large has larger number of trainable parameters than BERT-base, resulting from larger number of transformer encoder blocks and size of hidden dimension. The exact numbers of the parameters are in Appendix \ref{Appendix: Model architecture}. The pretrained checkpoints of BERT are downloaded from Huggingface \cite{wolf-etal-2020-transformers}. The names of the checkpoints for BERT-base and BERT-large are \verb|bert-base-uncased| and \verb|bert-large-uncased|, respectively. 

\begin{figure}[!hbt]
\centering
\setlength{\belowcaptionskip}{15pt}
\includegraphics[width=0.8\linewidth]{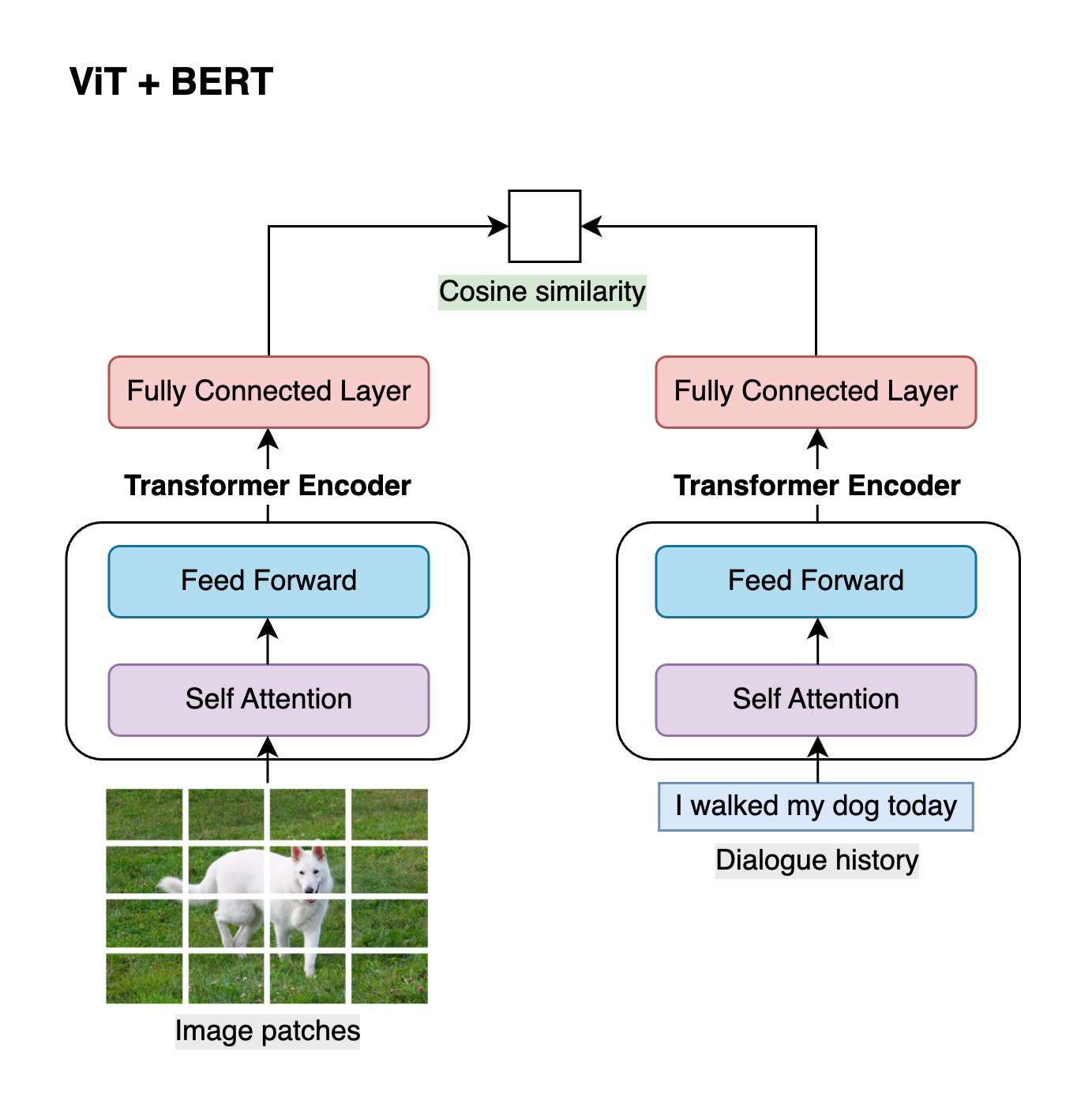}
\caption{Image retriever with ViT+BERT architecture.}
\label{fig:vit_plus_bert}
\end{figure}

An illustration of the complete image retriever using ViT+BERT architecture is provided in Figure \ref{fig:vit_plus_bert}. Each rounded rectangle represents each transformer block for simplicity, while the actual architectures of ViT and BERT consist of multiple blocks.

Image retriever is implemented in \verb|multimodal_chat/model/|\allowbreak\verb|image_retriever.py| of the source code. It contains \verb|class ImageRetriever()| that loads relevant tokenizer for text input, processor for image input, and model specified in the arguments to the main script. It also includes methods to load all candidate images and to retrieve the top 1 image using a finetuned model.

\subsection{Input and Output} \label{Implementation: Image retriever: Input and output}

As input to the image encoder, each image is rescaled to $224 \times 224$ for ViT-base or $384 \times 384$ for ViT-large. As input to the text encoder, each utterance of the dialogue history is concatenated using a separator token \verb|[SEP]|. In order to handle sequences of different lengths in minibatch training, each concatenated sequence is either padded to the right with padding token \verb|[PAD]| or truncated from right with maximum length of 512. 

The implementation of input and output is contained in \verb|class ImageRetriever|\allowbreak\verb|Collator()| of \verb|multimodal_chat/dataset/collator.py| of the source code. This class parses each dialogue session into dialogue history and target response for the text encoder. For a dialogue session with $n$ utterances, a total of $n-1$ pairs of dialogue history and target response are created, since any utterance besides the first one can be treated as a response to its previous utterances. This class also loads image from each URL into a PyTorch tensor of pixel values, which is processed into 3 RGB channels and is size normalized. The images are loaded during data collation instead of data preprocessing because loading thousands of images at once during processing causes CPU overload. After experimenting with various ways of storing images, it was found most efficient to store a mapping of each dialogue session and each image URL in the preprocessing step, and load the corresponding image in the collation step. This requires only a batch size number of images to be loaded at once.

\subsection{Training and Inference} \label{Implementation: Image retriever: Training and inference}

The training and inference of image retrievers require different implementation. During training, the image retriever only sees a batch size number of pairs of image and dialogue history. Among these pairs, the original image-dialogue is a positive pair, and all other combinations of image-dialogue are regarded as negative pairs. Thus, if the batch size is $bs$, there exist $1$ positive sample and $bs-1$ in-batch negative samples for each target sample. The existence of such positive and negative samples allows the use of contrastive loss function as the learning objective. Contrastive loss minimizes the distance between the positive pairs of image and dialogue history representations, while maximizing the distance between the negative pairs of image and dialogue history representations.

During inference, on the other hand, the image retriever gets access to all images in the test set, as it needs to rank all the image candidates to select the top 1 image. Given each dialogue history, the image retriever computes cosine similarity between the dialogue history representation and each image representation in the test set. This computation has linear time complexity, because if there are $n$ candidate images, the image encoder of image retriever needs $n$ forward passes to obtain $n$ representations and $n$ dot products to compute cosine similarity. Because $n$ forward passes cause extremely high latency for a chatbot model that needs to respond to each user in a few seconds, it is much more efficient to get representations of all candidate images prior to testing or deployment. Thus, in \verb|multimodal_chat/model/image_retriever.py|, the function \verb|load_images()| processes all image candidates into a tuple of (image representation, image URL) pairs. This function is called prior to test time, so that the computations required at real-time reduce to $n$ dot product operations and loading of the top 1 image from the mapped URL.

Moreover, at inference, after computing cosine similarity for all images in the database, a threshold logic controls the output. If the score of the image with maximum similarity is above a predefined threshold, the image is retrieved and shared to the user. If it is below the threshold, no image is retrieved. The threshold is empirically set as 0.15 from preliminary experiments on the validation set. Although this threshold does not affect any model performance, it determines the frequency with which the images are shared when a chatbot is deployed. The threshold should be between 0 and 1, and higher threshold results in fewer sharing of images in the conversation.

The training and evaluation batch sizes of image retrievers are both 16, which showed stable performance in preliminary experiments. The learning rate for all models is initially set to $5 \times 10^{-5}$ and decays over the steps with linear scheduling. The optimizer is AdamW \cite{loshchilov2019decoupled}, which adjusts the learning rate at each step using momentum and scaling. All models are trained for 10 epochs. 

The trainer class is for image retrievers implemented as \verb|class ImageRetrieverTrainer()| in \verb|multimodal_chat/|\allowbreak\verb|learning/trainer.py|. Inside this class, the main training loop is executed by the Huggingface \verb|Trainer| object, which automates training, evaluating, and saving models when dataset, model, and metrics are properly passed as arguments.

The main script is \verb|multimodal_chat/run_image_retriever.py|, in which all classes of \verb|ImageRetriever()|, \verb|ImageRetrieverCollator()|, and \verb|ImageRetrieverTrainer()| are loaded to perform training and evaluation of any variant of image retriever models.

\subsection{Computational Resource} \label{Implementation: Image retriever: Computational resource}

To train and evaluate all image retrievers, 1 NVIDIA TITAN RTX is utilized in a single-GPU setting. The training batch size is adjusted depending on the model size using \verb|per_device_|\allowbreak\verb|train_batch_size| and \verb|gradient_accumulation_steps| of arguments to Huggingface \verb|Trainer|. Because some model variants are too large and thus cause out of GPU memory when trained directly with batch size of 16, the actual batch size is set to some power of 2, and gradient accumulation is used to update the model weights only when 16 samples are seen. Thus, $bs = $ \verb|per_device_train_batch_size| $\times$ \verb|gradient_accumulation_steps|. In this way, all image retriever variants regardless of their number of parameters are trained with batch size of 16, thus removing variance of batch size when comparing their performance.

\section{Response Generator} \label{Implementation: Response Generator}

This section describes the implementation details of the response generator, including model architecture, input and output, training and inference, and computational resource.

\subsection{Unimodal Model Architecture} \label{Implementation: Response generator: unimodal model architecture}

For response generators that take in text input and return text output, the family of models used as response generators is GPT-2 \cite{radford2019language}. GPT-2 has a transformer decoder architecture and is pretrained on a massive corpus of web text. GPT-2 is widely used in text generation tasks such as summarization, question answering, story generation, and many more. Even though its generative performance was outperformed by billion-scale models such as GPT-3 \cite{https://doi.org/10.48550/arxiv.2005.14165}, GPT-2 still remains very competitive among million-scale models that are trainable with single GPU.

One line of GPT-2 is DialoGPT \cite{zhang-etal-2020-dialogpt}, which follows GPT-2 architecture but is additionally pretrained on various open-domain dialogue datasets. DialoGPT is reported to have better fluency in dialogue related tasks, thus suitable for a response generator.

GPT-2 generates text output given text input. For the purpose of response generation, the input and output are dialogue history and target response, respectively. Each transformer decoder block of GPT-2 consists of a masked self-attention layer and a feedforward layer. The purpose of masked self-attention layer is to create a contextual representation of each target token by attending to relevant tokens in dialogue history and current target tokens with different attention weights. The feedforward layer then maps this representation to a hidden representation space using both linear and nonlinear operations. The token representation returned from the last transformer decoder block is finally mapped to a vocabulary space $\mathbb{R}^V$ and processed with softmax function. The resulting vector in $\mathbb{R}^V$ is interpreted as a probability distribution over the target token, and a token sampled from this distribution becomes the generated token.

Two variants of GPT-2 are used in the experiment: GPT2-medium and DialoGPT-medium. The exact numbers of the parameters are in Appendix \ref{Appendix: Model architecture}. The pretrained checkpoints of these models are downloaded from Huggingface \cite{wolf-etal-2020-transformers}. The names of the checkpoints for GPT2-medium and DialoGPT-medium are \verb|gpt2-medium| and \verb|microsoft/|\allowbreak\verb|DialoGPT-medium|, respectively.

%These variants follows the basic GPT-2 architecture, except that they have different number of parameters rising from different number of transformer blocks and hidden dimension sizes. The exact numbers are in Appendix \ref{Appendix: Model architecture}.

\subsection{Multimodal Model Architecture} \label{Implementation: Response generator: multimodal model architecture}

For response generators that take in both text and image inputs and return text output, the model architecture used as response generators is \verb|VisionEncoderDecoder| from Huggingface \cite{wolf-etal-2020-transformers}. \verb|VisionEncoderDecoder| is a general class for an image encoder and a text decoder. Similarly to \verb|VisionTextEncoder| for implementing image retriever, \verb|VisionEncoderDecoder| is selected because any previously released image encoders and text decoders can be plugged into this class, allowing a lot of freedom in choice of models from each modality.

Prior to using \verb|VisionEncoderDecoder|, preliminary experiments were conducted using public multimodal encoder-decoder architectures including OFA \cite{wang2022ofa} and BLIP \cite{https://doi.org/10.48550/arxiv.2201.12086}. However, their performance on response generation was too low due to the relatively small size of text decoders around 110M parameters. This size was suitable for the target tasks of OFA and BLIP, which include image-text retrieval, image captioning, and visual question answering that require generation of only a few tokens such as captions and short answers (e.g., ``How many dogs are in this photo?" with the answer ``1"). On the other hand, the response generator in my chatbot system requires a sufficiently large text decoder capable of generating longer sequences. Thus, \verb|VisionEncoderDecoder| was a better option as it allows the use of any pretrained text decoders regardless of their size.

%For response generators that take in both text and image input and output text, the family of models used in the experiment is BLIP \cite{https://doi.org/10.48550/arxiv.2201.12086}. BLIP is a vision-text transformer-based model pretrained on massive web data of images and their captions and descriptions. BLIP demonstrates competitive performance on various mutlimodal tasks including image-text retrieval, image captioning, and visual question answering.

%Among other architectures of vision-text model, BLIP is chosen for multiple reasons. First, the main component of BLIP is also a transformer block similar to GPT-2, so comparison between GPT-2 and BLIP architecture reduces variance and allows relatively fair analysis of existence of additional image input to response generators. Second, BLIP performed well on visual question answering which is most analogous to the task of response generation, as the dialogue history can roughly be thought of as a question from the user that needs to be responded. Third, it is well implemented in Huggingface.

\verb|VisionEncoderDecoder| consists of two main architectural components: an image encoder and a text decoder. For the task of response generation, the image encoder takes in either the currently retrieved or the most recently retrieved image and encodes it into an image representation. Then, the text decoder takes in this image representation and combines it with representations of dialogue history and current response tokens to generate the next response token. 

The family of models used as image encoders in multimodal response generators is ViT \cite{DBLP:journals/corr/abs-2010-11929}, which is also used as image encoders in image retrievers. In ViT, a sequence of image patches is encoded into an image representation using self attention and feedforward layers. The architecture of ViT is explained in Section \ref{Implementation: Image retriever: Model architecture} in further detail.

Two variants of ViT are used in the experiment: ViT-base and ViT-large. The pretrained checkpoints from Huggingface \cite{wolf-etal-2020-transformers} for ViT-base and ViT-large are \verb|google/vit-base-|\allowbreak\verb|patch16-|\allowbreak\verb|224| and \verb|google/vit-large-patch32-384| respectively, same as the image retrievers in Section \ref{Implementation: Image retriever: Model architecture}.

On the other hand, the family of models used as text decoders in multimodal response generators is GPT-2 \cite{radford2019language}, which is also experimented in a unimodal setting. In GPT-2, the dialogue history and current response tokens are encoded a text representation using masked self attention. This text representation then attends over the relevant parts of the image representation returned from the image encoder through cross attention. This cross attention layer is not part of the pretrained GPT-2, thus trained from scratch during finetuning. The resulting representation goes through feedforward and embedding layers, finally sampled as a generated token. The architecture of GPT-2 is explained in Section \ref{Implementation: Response generator: unimodal model architecture} in further detail.

Two variants of GPT-2 are used in the experiment: GPT2-medium and DialoGPT-medium. The pretrained checkpoints from Huggingface \cite{wolf-etal-2020-transformers} for GPT2-medium and DialoGPT-medium are \verb|gpt2-medium| and \verb|microsoft/DialoGPT-medium| respectively, same as the unimodal response generator in Section \ref{Implementation: Response generator: unimodal model architecture}.

%BLIP for question answering consists of three architectural components: image encoder, question encoder, and answer decoder. For the task of response generation, the image encoder takes in the image shared in the dialogue and encode into a representation using self-attention. Then, the question encoder takes in the textual dialogue history and combine its representation with the image presentation to a joint representation space by cross-attention, to create the final image-text input encoding. Finally, the answer decoder takes in this image-text encoding and combines with encoding of previous response tokens through cross-attention to generate the next response token by token. All encoder and decoder use feedforward network after attention in order to map the encoding to either hidden or decoding space.

\begin{figure}[!hbt]
\centering
\setlength{\belowcaptionskip}{15pt}
\includegraphics[width=1.0\linewidth]{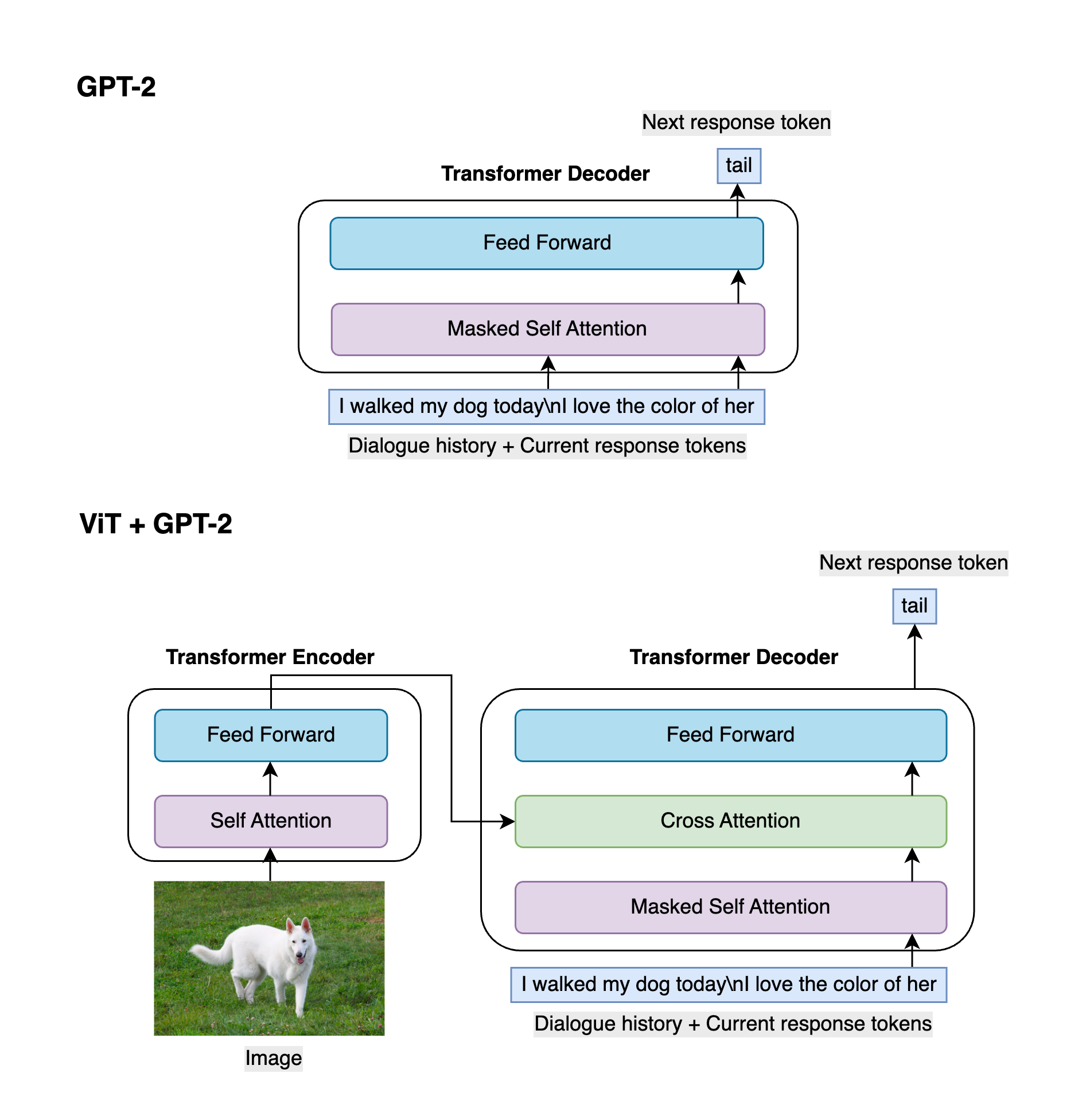}
\caption{Comparison of GPT-2 and ViT+GPT-2 architectures.}
\label{fig:gpt2_and_vit_plus_gpt2}
\end{figure}

The difference between GPT-2 as unimodal response generator and ViT+GPT-2 as multimodal response generator is illustrated in Figure \ref{fig:gpt2_and_vit_plus_gpt2}. Each rounded rectangle represents each transformer block for simplicity, while the actual architectures of GPT-2 and ViT+GPT-2 consist of multiple blocks.

%Two variants of BLIP are used in the experiment: BLIP-base and BLIP-large. The two variants follow the basic BLIP architecture, except that BLIP-large has more trainable parameters and is additionally trained on self-supervised dataset, in which a caption model generate pseudo captions for unlabeled image dataset and a filter model filters image-caption pairs by their quality. The exact numbers are in Appendix \ref{Appendix: Model architecture}.

Response generator is implemented in \verb|multimodal_chat/model/response_generator.py| of the source code. It contains \verb|class ResponseGenerator()| that loads relevant tokenizer for text input, processor for image input if applicable, and model specified in the arguments to the main script. It also includes a method to properly infer a finetuned response generator, either unimodal or multimodal.

\subsection{Input and Output} \label{Implementation: Response generator: Input and output}

As input to the image encoder, each image is rescaled to $224 \times 224$ for ViT-base or $384 \times 384$ for ViT-large. As input to the text decoder, dialogue history and response should be accurately processed in order to adhere to Huggingface model input and output convention. First, the vanilla version of GPT-2 computes loss over all tokens given as labels; however, for the purpose of response generation, GPT-2 needs to compute loss only over the portion of the response, using dialogue history merely as a condition. Thus, labels are padded from left with dummy token \verb|-100| with length of dialogue history, which is ignored by the model when computing loss. Moreover, five special tokens are added to GPT-2 tokenizer: \verb|<user>|, \verb|<bot>|, \verb|<bos>|, \verb|<eos>|, and \verb|<pad>|. Each user or bot utterance in the dialogue session is prepended with \verb|<user>| or \verb|<bot>| respectively and concatenated into a single sequence. The resulting sequence is then prepended with \verb|<bos>| and appended with \verb|<eos>|. Shorter sequences are padded with \verb|<pad>| on the right while longer sequences are truncated from the right, in order to support multi-batch training. An example is given in Table \ref{table:gpt2_input_and_output}. Here, \verb|input_ids| is the entire concatenated sequence of dialogue history and target response, and \verb|labels| is the portion of the sequence in which loss is computed. \verb|input_ids| is shifted one position to the right automatically inside the model for proper alignment of input and output.

\begin{table}[!hbt]
  \centering
  \vspace*{15pt}
  \begin{tabular}{l|r}
   \hline \hline
    \verb|input_ids| & \verb!<bos> <user> I walked my dog today <bot> she's adorable <eos>! \\
    \hline
    \verb|labels| & \verb!-100 -100 -100 -100 -100 -100 -100 -100 she's adorable <eos>! \\
    \hline \hline
  \end{tabular}
  \caption{Input and output format of GPT-2 for response generation.}
  \label{table:gpt2_input_and_output}
\end{table}

%For BLIP variants, the architecture dialogue history and target response are simply given as inputs and labels without any left padding. 

The implementation of input and output is contained in \verb|class ResponseGenerator|\allowbreak\verb|Collator()| of \verb|multimodal_chat/dataset/collator.py| of the source code. This class parses each dialogue session into dialogue history and target response and make them into batched tensors with relevant padding and masking. Similarly to \verb|class ImageRetriever|\allowbreak\verb|Collator()|, given a dialogue session with $n$ utterances, a total of $n-1$ pairs of dialogue history and target response are created. Additionally for multimodal response generators, this class loads image from each URL into a PyTorch tensor of pixel values, which is processed into 3 RGB channels and normalized into size of $224 \times 224$ for ViT-base or $384 \times 384$ for ViT-large. Loading images during data collation is more efficient than loading them during data preprocessing, as explained in Section \ref{Implementation: Image retriever: Input and output}.

\subsection{Training and Inference} \label{Implementation: Response generator: Training and inference}

The learning objective of response generation is implemented by the cross entropy function. When predicting each token in the target response, the true distribution of the token is given as a one-hot vector in the vocabulary space $\mathbb{R}^V$, in which the value at the index of the target token is 1 and the others are 0. The predicted distribution of the token is the output of the final embedding layer mapping to $\mathbb{R}^V$, followed by a softmax function to make the output into a probability distribution. Thus, minimizing the cross entropy between these two distributions is equivalent to making the predicted distribution more like the true distribution. A single loss term is computed for each target token in this way, and the average of these loss terms over all tokens in the target response become the final loss for each sample.

The training and inference of response generators are different in terms of the availability of the target response in predicting each target token. Training occurs in a teacher-forcing manner, meaning that when predicting the current target token, all previous target tokens are given. Thus, even if the current prediction is wrong, this error does not propagate to subsequent tokens, which make predictions independently using the target tokens prior to that point. On the other hand, inference has no access to target tokens. Only the dialogue history is given as the input, and once the model generates a token, this token gets concatenated to the dialogue history for all subsequent tokens.

Moreover, at the inference of multimodal response generators, no image is available at current inference step if the similarity score of the top 1 image does not pass the threshold. In this case, the most recently retrieved image in the dialogue is fed to the image encoder of the response generator instead. This is a reasonable implementation since the response generator still needs to understand the image previously shared in the dialogue in order to generate consistent responses. A few edge cases exist: if there are multiple previously shared images, only the most recent one is given as the image input, based on a heuristic that the user will talk mostly about the recent image. If there are no previously shared images, then a dummy image of zero pixels is given as the image input, and the response generator is expected to ignore this image when generating responses.

The training and evaluation batch sizes of response generators are 16 and 4 respectively, which showed stable performance in preliminary experiments. The evaluation batch size is smaller than the training batch size because generating responses at each evaluation step requires attention operation with quadratic time complexity, which consumes significant amount of GPU memory and thus sensitive to batch size. The learning rate for all models is initially set to $5 \times 10^{-5}$ and decays over the steps with linear scheduling. The optimizer is AdamW \cite{loshchilov2019decoupled}, which adjusts the learning rate at each step using momentum and scaling. All models are trained for 3 epochs. 

The trainer class for response generators is implemented as \verb|class ResponseGenerator|\allowbreak\verb|Trainer()| in \verb|multimodal_chat/|\allowbreak\verb|learning/trainer.py|. Inside this class, the main training loop is executed by the Huggingface \verb|Trainer| object, as in training of image retrievers. Custom metrics and postprocessing steps are separately implemented and passed to \verb|Trainer| via function callback. The callback class is \verb|class ResponseGeneratorCallback()| in \verb|multimodal_chat/|\allowbreak\verb|learning/|\allowbreak\verb|callback.py|, which computes relevant metrics and send their results to Tensorboard for monitoring and evaluation.

The main script is \verb|multimodal_chat/run_response_generator.py|, in which all classes of \verb|ResponseGenerator()|, \verb|ResponseGeneratorCollator()|, and \verb|ResponseGenerator|\allowbreak\verb|Trainer()| are loaded to perform training and evaluation of any variant of response generator models.

\subsection{Computational Resource} \label{Implementation: Response generator: Computational resource}

To train and evaluate all response generators, 1 NVIDIA TITAN RTX is utilized in a single-GPU setting. Similarly to the case of image retrievers, the training batch size is adjusted depending on the model size using \verb|per_device_train_batch_size| and \verb|gradient_|\allowbreak\verb|accumulation_|\allowbreak\verb|steps| of arguments to Huggingface \verb|Trainer|.

\chapter{Evaluation}
In this section, the proposed system is evaluated using various evaluation methods. First, image retrievers and response generators are independently evaluated through automatic evaluation in Sections \ref{Automatic Evaluation: Image Retriever} and \ref{Automatic Evaluation: Response Generator}, respectively.
Then, the complete chatbot system is evaluated through human evaluation in Section \ref{Human Evaluation: Chatbot}.

\section{Dataset} \label{Evaluation: Dataset}

\subsection{Dataset Description and Statistics} \label{Evaluation: Data description and statistics}

\begin{figure}[!hbt]
\centering
\vspace*{15pt}
\setlength{\belowcaptionskip}{15pt}
\includegraphics[width=0.9\linewidth]{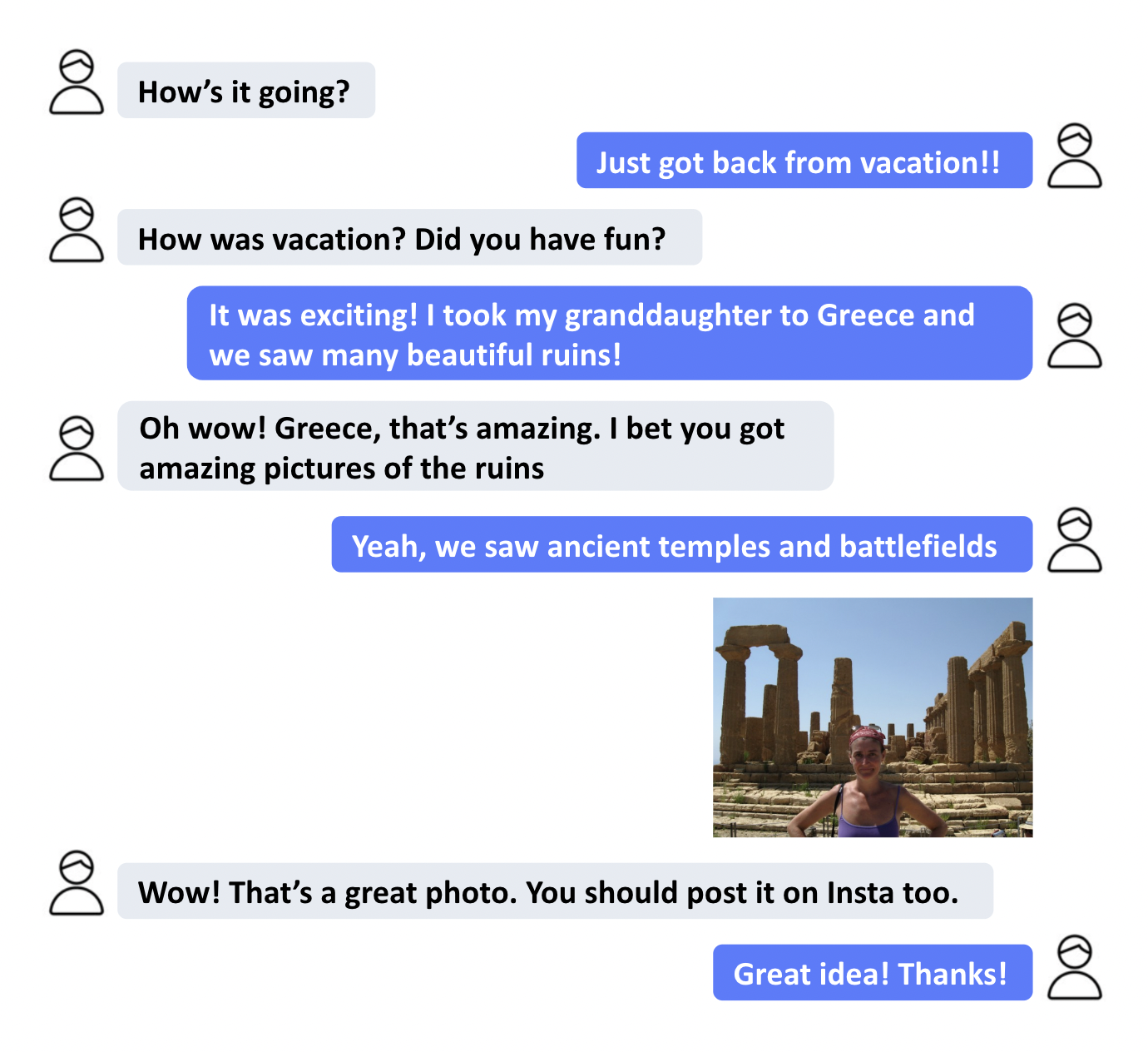}
\caption{Example dialogue from PhotoChat dataset.}
\label{fig:photochat_sample}
\end{figure}

The dataset used to train and evaluate all models is PhotoChat \cite{zang-etal-2021-photochat}, an open-domain dialogue dataset in which two speakers converse over approximately 12 turns and an image is shared at some point during the conversation. An example of a dialogue in PhotoChat is in Figure \ref{fig:photochat_sample}. PhotoChat was collected by crowdworkers, where two crowdworkers are instructed to talk to each other about any topic as if they are talking to their friend. In each conversation, one of the crowdworkers has access to a randomly selected image from a database of images, and this crowdworker is additionally instructed to drive the conversation such that they can share the given image at appropriate time. The database of images is a filtered subset of Open Image Dataset V4 (OID) \cite{Kuznetsova_2020}, containing images only with objects commonly shared in daily conversations (people, food, animal, and product). PhotoChat is publicly released to facilitate research on multimodal modeling of image and dialogue.

The statistics of PhotoChat is shown in Table \ref{table:photochat_statistics}. There are approximately 12,000 pairs of image and dialogue. Same image may be paired with multiple dialogues, so the total number of images is less than the total number of dialogues. Each dialogue contains approximately 12.7 turns and 80.4 tokens on average.

\begin{table}[!hbt]
  \centering
  \setlength{\belowcaptionskip}{15pt}
  \begin{tabular}{l|r|r|r|r}
   \hline \hline
    \textbf{Split} & \textbf{\# Images}& \textbf{\# Dialogues} & \textbf{\# Turns} & \textbf{\#  Tokens} \\
    \hline \hline
    Train & 8,917 & 10,286 & 130,546 & 827,154 \\
    \hline
    Validation & 1,000 & 1,000 & 12,701 & 80,214 \\
    \hline
    Test & 1,000 & 1,000 & 12,852 & 80,847 \\
    \hline
    \hline
    Total & 10,917 & 12,286 & 156,099 & 988,215 \\
    \hline \hline
  \end{tabular}
  \caption{Statistics of original PhotoChat dataset.}
  \label{table:photochat_statistics}
\end{table}

\subsection{Dataset Preprocessing} \label{Evaluation: Data preprocessing}

In order to train and evaluate the proposed image retrievers and response generators, the original PhotoChat \cite{zang-etal-2021-photochat} is preprocessed in multiple steps.

First, in each pair of image and dialogue in PhotoChat, the image is stored not as pixels but as URL to its remote location. This location is different for all images, as the original OID \cite{Kuznetsova_2020} is a collection of images from various sources on web. Thus, the URLs of some images have expired since PhotoChat was released, and the images are no longer available. These images are excluded from the final dataset.

Second, in each dialogue in PhotoChat, an image is shared once, which means that only one turn is explicitly paired with the image and all other turns are text only. This imbalance is not an issue for unimodal generators, which do not use any images as model input. However, for multimodal generators, the turns without any paired image cannot be used in its original format. As a solution, all turns after the shared image are paired with the same image, which is considered as the most recently shared image in the dialogue. Also, all turns before the shared image are paired with a dummy image of zero pixels, expected to be ignored by the response generator. With this preprocessing step, all turns of each dialogue can be used by both types of response generators. One possible alternative would be simply not to use the turns without images, but this would decrease the preprocessed dataset size to about 10\% of the original size, which is a very ineffective use of the dataset. Another alternative is to use an ensemble of a unimodal generator and a multimodal generator, in which the unimodal generator is trained with full dataset and the multimodal generator is trained only with turns paired with images. However, this method would cost twice the GPU memory at both training and inference time, so it was not a feasible option. 

%This is found to be the most efficient solution to this imbalanced dataset, since at inference when there is no shared image, the same black dummy image can be fed into the response generator without additional computational cost.

Third, each dialogue in PhotoChat may not consist of perfectly alternating turns between the two speakers. One speaker may say multiple utterances before the opposite speaker says anything, just like in real life. This occurs in about 25\% of dialogues in PhotoChat. However, to simplify the task of response generation, the consecutive utterances from the same speaker are concatenated with a whitespace into a single sequence. A qualitative observation over PhotoChat supports that this preprocessing is appropriate in most cases, as the multiple utterances are usually segments of a full grammatical sentence.

Lastly, in some dialogues in PhotoChat, a speaker responds to the opposite speaker without any text but with image only. Although this may happen in real life, the proposed chatbot system is intended to respond to user with some text at every turn. Thus, for these dialogues, the image is instead paired to the subsequent utterance of the same speaker. This makes all images explicitly paired to text, allowing image retrievers and response generators to be trained with full dataset without any unused dialogues.

All preprocessing code is included in \verb|multimodal_chat/dataset/processor.py|.

\subsection{Final Dataset}

Table \ref{table:preprocessed_dataset} shows the final dataset after preprocessing PhotoChat \cite{zang-etal-2021-photochat}. This dataset is used to train, validate, and test all image retrievers and response generators in Sections \ref{Automatic Evaluation: Image Retriever} and \ref{Automatic Evaluation: Response Generator}. All models use the same split from the original paper. The train split is used for training the models, the validation split is used for selecting best model checkpoints, and the test split is used for evaluating the performance of the best model checkpoint. 

\begin{table}[!hbt]
  \centering
  \setlength{\belowcaptionskip}{15pt}
  \begin{tabular}{l|r|r}
   \hline \hline
    \textbf{Split} & \textbf{\# Samples for image retriever} & \textbf{\# Samples for response generator} \\
    \hline \hline
    Train & 10,286 & 89,402 \\
    \hline
    Validation & 1,000 & 8,751 \\
    \hline
    Test & 1,000 & 8,776 \\
    \hline
    \hline
    Total & 12,286 & 106,929 \\
    \hline \hline
  \end{tabular}
  \caption{Number of PhotoChat samples after preprocessing. For image retrievers, each sample is a pair of (image, dialogue history). For response generators, each sample is a triple of (image, dialogue history, response).}
  \label{table:preprocessed_dataset}
\end{table}

%For image retriever, the train and validation sets contain only valid images, but the test set contains both valid and dummy images. This is because at training, the model learns to retrieve correct images, but at inference, the model also has to distinguish whether to retrieve any image or not given dialogue history. 

\section{Automatic Evaluation: Image Retriever} \label{Automatic Evaluation: Image Retriever}

\subsection{Models}

Four types of image retrievers are trained and evaluated for comparison of performance. The models are in the order of increasing number of parameters. 

\begin{itemize}
\item \textit{ViT-base+BERT-base}: This model has ViT-base \cite{DBLP:journals/corr/abs-2010-11929} as image encoder and BERT-base \cite{devlin-etal-2019-bert} text encoder. It has approximately 196M trainable parameters, with 86M parameters from ViT-base and 110M parameters from BERT-base. The details of its architecture and format of input and output are explained in Sections \ref{Implementation: Image retriever: Model architecture} and \ref{Implementation: Image retriever: Input and output}.
\item \textit{ViT-large+BERT-base}: This model has ViT-large \cite{DBLP:journals/corr/abs-2010-11929} as image encoder and BERT-base \cite{devlin-etal-2019-bert} as text encoder. It has approximately 417M trainable parameters, with 307M parameters from ViT-large and 110M parameters from BERT-base. The architecture and format of input and output are same as \textit{ViT-base+BERT-base}.
\item \textit{ViT-base+BERT-large}: This model has ViT-base \cite{DBLP:journals/corr/abs-2010-11929} as image encoder and BERT-large \cite{devlin-etal-2019-bert} as text encoder. It has approximately 422M trainable parameters, with 86M parameters from ViT-base and 336M parameters from BERT-large. The architecture and format of input and output are same as \textit{ViT-base+BERT-base}.
\item \textit{ViT-large+BERT-large}: This model has ViT-large \cite{DBLP:journals/corr/abs-2010-11929} as image encoder and BERT-large \cite{devlin-etal-2019-bert} as text encoder. It has approximately 643M trainable parameters, with 307M parameters from ViT-large and 336M parameters from BERT-large. The architecture and format of input and output are same as \textit{ViT-base+BERT-base}.
\end{itemize}

\subsection{Metrics}

The evaluation metrics for image retrievers are Recall@K and Mean Reciprocal Rank (MRR).
%are image-sharing classification F1 and Recall@K.

\begin{itemize}
    %\item Image-sharing classification F1: 
    \item \textit{Recall@K}: This metric measures whether the gold image is among the highly ranked images. If the gold image exists in the top $K$ ranked images, the score of $Recall@K$ is 1. If it does not exist in the top $K$ ranked images, the score of $Recall@K$ is 0. If $K_1 \leq K_2$, it follows that $Recall@K_1 \leq Recall@K_2$.
    \item \textit{Mean Reciprocal Rank (MRR)}: This metric measures how high the gold image is ranked out of all candidate images. If the rank of gold image is $r$, the score of MRR is $\frac{1}{r}$.
\end{itemize}

\subsection{Main Results}

Table \ref{table:image-retriever} shows the performance of each image retriever on the test set of PhotoChat \cite{zang-etal-2021-photochat}. The best checkpoint for each image retriever is chosen by minimum loss over the validation set. Additionally, two image retrievers with best performance from the original PhotoChat paper are included for comparison with the four proposed image retrievers. \textit{VSE++} \cite{faghri2018vse} has a dual-encoder architecture with ResNet152 \cite{he2015deep} as image encoder and GRU \cite{cho-etal-2014-learning} as text encoder. \textit{SCAN} \cite{lee2018stacked} uses cross-attention architecture between image region embedding from Faster R-CNN \cite{ren2016faster} and text embedding from GRU \cite{cho-etal-2014-learning}. The scores of Recall@1/5/10 for \textit{VSE++} and \textit{SCAN} are copied from the PhotoChat paper, and MRR is not reported. All six models are trained and evaluated with the same dataset split in Table \ref{table:photochat_statistics}.

\begin{table}[!hbt]
  \centering
  \setlength{\belowcaptionskip}{15pt}
  \begin{tabular}{l|ccc}
   \hline \hline
    \textbf{Model} & \textbf{Recall@1/5/10$\uparrow$} & \textbf{MRR$\uparrow$}\\
    \hline \hline
    \textit{VSE++} & 0.102/0.254/0.342 & - \\
    \hline
    \textit{SCAN} & \textbf{0.104}/0.270/0.371 & - \\
    \hline \hline
    \textit{ViT-base+BERT-base} & \textbf{0.104}/0.304/\textbf{0.436} & \textbf{0.212} \\
    \hline
    \textit{ViT-large+BERT-base} & 0.097/\textbf{0.312}/0.427 & 0.210 \\
    \hline
    \textit{ViT-base+BERT-large} & 0.093/0.282/0.424 & 0.200 \\
    \hline
    \textit{ViT-large+BERT-large} & 0.092/0.287/0.406 &0.202 \\
    \hline \hline
  \end{tabular}
  \caption{Automatic evaluation results of various image retrievers on the test set of PhotoChat. The best score for each metric is in bold.}
  \label{table:image-retriever}
\end{table}

First, in terms of Recall@1/5/10, all four proposed models achieve approximately 0.1/0.3/0.4. This means that the ground-truth image is included in the top 1 rank 10\% of the times, in the top 5 rank 30\% of the times, and in the top 10 rank 40\% of the times on average. This is a significant performance given that the total number of candidate images is 1,000 as mentioned in Table \ref{table:preprocessed_dataset}. All four models achieve higher Recall@5/10 than \textit{VSE++} and \textit{SCAN} with maximum difference of +0.06/0.09 and +0.04/0.06, respectively, which demonstrates that transformer-based dual encoder architecture is effective in modeling image retrievers, even outperforming cross encoder architecture. Moreover, among the four proposed models, Recall@1/10 is the highest in \textit{ViT-base+BERT-base} by a small margin. This is a counter-intuitive finding, since larger models are usually expected to perform better. It can be conjectured that the mapping of image and dialogue history is explicit enough in PhotoChat such that small encoders are sufficient for the task of image retrieval.

Furthermore, in terms of MRR, all four proposed models achieve approximately 0.2. This means that the ground-truth image is ranked as the top 5 image on average, implying that it is typically ranked higher than the other 995 images. Among the four proposed models, \textit{ViT-base+BERT-base} and \textit{ViT-large+BERT-base} score slightly higher than \textit{ViT-base+BERT-large} and \textit{ViT-large+BERT-large} by a margin of +0.01 in MRR. This shows that BERT-base is sufficiently competitive compared to BERT-large as text encoders in image retrieval, despite being three times smaller in size.

\begin{figure}[!hbt]
\centering
\vspace*{15pt}
\setlength{\belowcaptionskip}{15pt}
\includegraphics[width=1.0\linewidth]{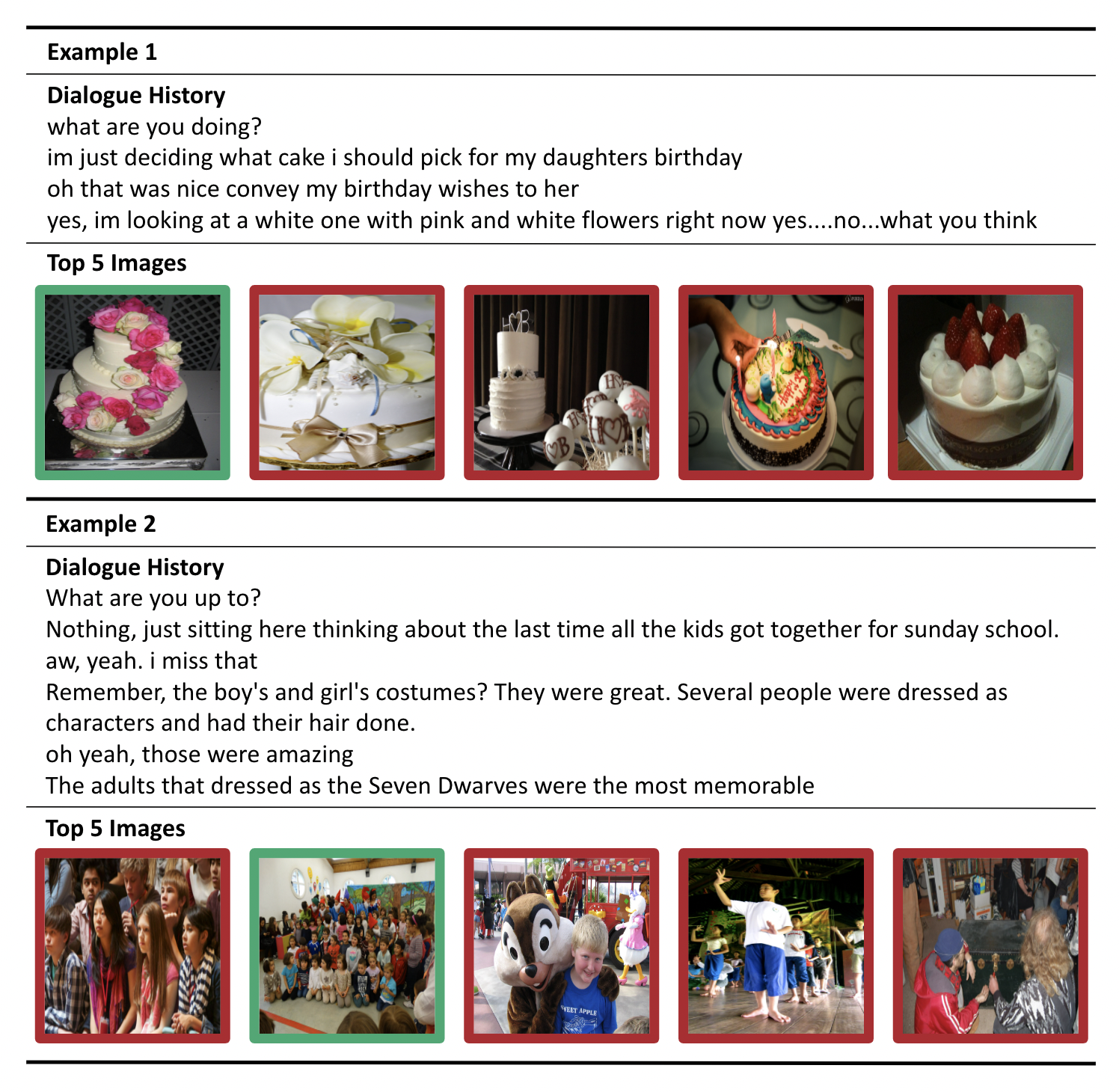}
\caption{Examples of top 5 images retrieved by \textit{ViT-base+BERT-base} given each dialogue history in the test set of PhotoChat. The order of images is from rank 1 (left) to rank 5 (right). The ground-truth image is in green, and other images are in red.}
\label{fig:sample_image_retriever}
\end{figure}

Figure \ref{fig:sample_image_retriever} shows the examples of top 5 images ranked by the image retriever given each dialogue history in the test set of PhotoChat. The model used to retrieve the images is \textit{ViT-base+BERT-base}. In Example 1, the two speakers are talking about getting a birthday cake with pink and white flowers. The image retriever understand the details of the cake mentioned in dialogue history and correctly retrieves the ground-truth image as top 1 image. The top 2-5 images are also images of cakes, but their color or shape is different from what is mentioned in the conversation, which the image retriever is able to distinguish. In Example 2, the two speakers are reminiscing about the time when their children dressed up in costumes at Sunday school. The ground-truth image is ranked as the top 2 image. Given the dialogue history, the image retriever looks for images with children in costumes, but these details become hardly visible as the images are scaled down to $224 \times 224$, thus reducing precision. The other top candidates are reasonably confusing to the model, such as the top 1 image with children but without any costumes or the top 3 image with a costume but only a single child.

\subsection{Visualization of Model Training}

\begin{figure}[!hbt]
\centering
\vspace*{15pt}
\setlength{\belowcaptionskip}{15pt}
\includegraphics[width=1.0\linewidth]{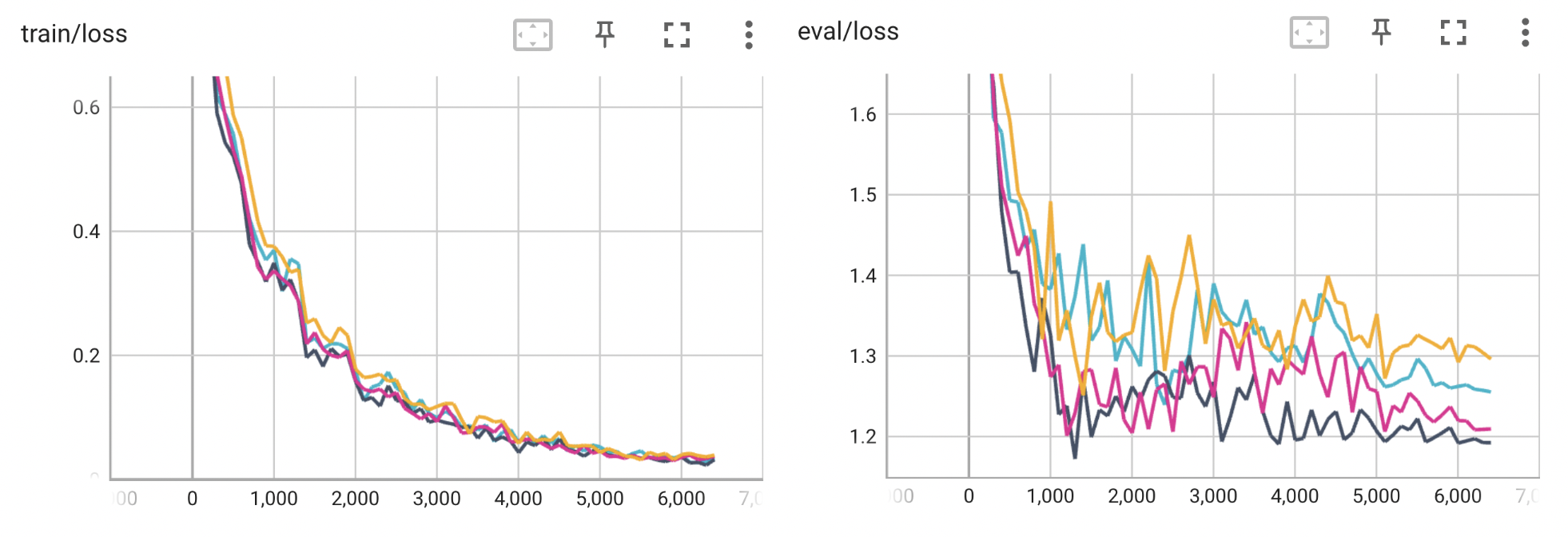}
\caption{Training and evaluation loss of image retrievers reported to TensorBoard. Each image retriever is \textit{ViT-base+BERT-base} (navy), \textit{ViT-large+BERT-base} (pink), \textit{ViT-base+BERT-large} (blue), and \textit{ViT-large+BERT-large} (yellow). Outliers are hidden from display.}
\label{fig:loss_image_retriever}
\end{figure}

The training and evaluation losses for each image retriever per training step are shown in Figure \ref{fig:loss_image_retriever}. All models are trained for 10 epochs, which amounts to approximately 6,400 steps in total. At every 100 steps, the training loss is computed by averaging the loss terms from the 100 training samples. Then, the model is frozen for evaluation and is fed with all samples in the validation set, whose loss terms are averaged to compute the evaluation loss. The automatic evaluation metrics Recall@K and MRR cannot be computed at every evaluation step, because computing these metrics require encoding of all candidate images with the checkpoint at that step, which takes excessive computational time.

On the left side of Figure \ref{fig:loss_image_retriever}, training loss decreases steadily for all four image retrievers over 10 epochs. This indicates that the contrastive loss function does proper weight update and the models are able to converge.

On the right side of Figure \ref{fig:loss_image_retriever}, the evaluation loss also decreases for all four image retrievers, but with some variance among the models. At the final step, the evaluation loss is in the increasing order of \textit{ViT-base+BERT-base}, \textit{ViT-large+BERT-base}, \textit{ViT-base+BERT-large}, and \textit{ViT-large+BERT-large} with largest gap of 0.1. Although small in absolute value, such gap still suggests that larger image retrievers may require more training epochs to fully converge than smaller image retrievers. 

\subsection{Source code}

Each image retriever is evaluated in \verb|class ImageRetrieverEvaluator()| of \verb|multimodal_chat/|\allowbreak\verb|learning/evaluator.py|. Here, the best model checkpoint is given as an argument to this class, and this model encodes all candidate images in the test set into image representations in advance. Then, each dialogue history in the test set is encoded into a representation, and all candidate images are sorted in terms of their similarity score with the dialogue history representation. This sorted order is used to compute Recall@K and MRR.

\section{Automatic Evaluation: Response Generator} \label{Automatic Evaluation: Response Generator}

\subsection{Models}

Six types of response generators are trained and evaluated for comparison of performance. The models are in the order of increasing number of parameters.

\begin{itemize}
    \item \textit{GPT2-medium}: This model has GPT2-medium \cite{radford2019language} as text decoder. It has a unimodal architecture. It has approximately 355M trainable parameters. The details of its architecture and format of input and output are explained in Sections \ref{Implementation: Response generator: unimodal model architecture} and \ref{Implementation: Response generator: Input and output}. 
    \item \textit{DialoGPT-medium}: This model has DialoGPT-medium \cite{zhang-etal-2020-dialogpt} as text decoder. It has a unimodal architecture. It has approximately 355M trainable parameters. The architecture and format of input and output are same as \textit{GPT2-medium}.
    \item \textit{ViT-base+GPT2-medium}: This model has ViT-base \cite{DBLP:journals/corr/abs-2010-11929} as image encoder and GPT2-medium \cite{radford2019language} as text decoder. It has a multimodal architecture. It has approximately 541M trainable parameters, with 86M parameters from ViT-base, 355M parameters from GPT2-medium, and 100M parameters from additional cross attention layers. The details of its architecture and format of input and output are explained in Sections \ref{Implementation: Response generator: multimodal model architecture} and \ref{Implementation: Response generator: Input and output}. 
    \item \textit{ViT-base+DialoGPT-medium}: This model has ViT-base \cite{DBLP:journals/corr/abs-2010-11929} as image encoder and DialoGPT-medium \cite{zhang-etal-2020-dialogpt} as text decoder. It has a multimodal architecture. It has approximately 541M trainable parameters, with 86M parameters from ViT-base, 355M parameters from DialoGPT-medium, and 100M parameters from additional cross attention layers. The architecture and format of input and output are same as \textit{ViT-base+GPT2-medium}.
    \item \textit{ViT-large+GPT2-medium}: This model has ViT-large \cite{DBLP:journals/corr/abs-2010-11929} as image encoder and GPT2-medium \cite{radford2019language} as text decoder. It has a multimodal architecture. It has approximately 762M trainable parameters, with 307M parameters from ViT-large, 355M parameters from GPT2-medium, and 100M parameters from additional cross attention layers. The architecture and format of input and output are same as \textit{ViT-base+GPT2-medium}.
    \item \textit{ViT-large+DialoGPT-medium}: This model has ViT-large \cite{DBLP:journals/corr/abs-2010-11929} as image encoder and DialoGPT-medium \cite{zhang-etal-2020-dialogpt} as text decoder. It has a multimodal architecture. It has approximately 762M trainable parameters, with 307M parameters from ViT-large, 355M parameters from DialoGPT-medium, and 100M parameters from additional cross attention layers. The architecture and format of input and output are same as \textit{ViT-base+GPT2-medium}.
    %\item \textit{GPT2-large}: This model is a GPT-2 variant with 774M trainable parameters.  \cite{radford2019language}. The architecture and format of input and output is same as \textit{GPT2-medium}. The only difference is that \textit{GPT2-large} has twice the number of parameters.
    %\item \textit{BLIP-base}: This model is a BLIP variant with 385M parameters \cite{https://doi.org/10.48550/arxiv.2201.12086}. The details of its architecture and format of input and output are in \ref{Implementation: Response generator: multimodal model architecture} and \ref{Implementation: Response generator: Input and output}. 
    %\item \textit{BLIP-large}: This model is a BLIP variant with 385M parameters \cite{https://doi.org/10.48550/arxiv.2201.12086}. The architecture is same as \textit{BLIP-base}.
\end{itemize}

\subsection{Metrics} \label{Automatic evaluation: response generator: metrics}

The evaluation metrics for response generators are Perplexity (PPL), BLEU-1/2 \cite{papineni-etal-2002-bleu}, and Distinct-1/2 \cite{li-etal-2016-diversity}.

\begin{itemize}
    \item \textit{Perplexity (PPL)}: This metric measures how probable the gold responses are. With respect to the validation or test sets, each finetuned model outputs probability over the gold response given each dialogue history. The difference between predicted and gold probabilities is computed via cross entropy, and its exponential is perplexity. 
    %This metric is  a measure of model fitting.
    %This metric is a measure of fluency, under the assumption that the model that assigns a higher probability to gold responses has learned a better language distribution.
    It is implemented by \verb|torch.nn.CrossEntropyLoss()|.
    \item \textit{BLEU-1/2} \cite{papineni-etal-2002-bleu}: This metric measures lexical similarity between generated response and gold response. With respect to the validation or test sets, each finetuned model generates a response given each dialogue history, and the fraction of n-gram overlaps between each generated response and gold response is computed. BLEU-1/2 computes 1/2-gram overlaps, respectively. It is implemented using \verb|nltk| library.
    %This metric is a measure of quality, under the assumption that the more similar the generated response is to the gold response, it had better quality.
    \item \textit{Distinct-1/2} \cite{li-etal-2016-diversity}: This metric measures diversity of the generated response. With respect to the validation or test sets, each finetuned model generates a response given each dialogue history, and the fraction of unique n-grams in each generated response is computed. Distinct-1/2 computes number of 1/2-grams, respectively. It is implemented using \verb|lexical-diversity| library.
    %This metric is a measure of diversity, under the assumption that the more unique tokens in generated response, it is more diverse.
\end{itemize}

The implementation of the above metrics are in \verb|multimodal_chat/util/metric.py|.

\subsection{Main Results}

Table \ref{table:response-generator} shows the performance of each response generator on the test set of PhotoChat. The best checkpoint for each response generator is chosen by minimum loss over the validation set. Additionally, a unimodal response generator proposed in a previous work \cite{sun-etal-2022-multimodal} is included for comparison with the six proposed models. This is the only existing work that implements a chatbot with photo sending abilities, as mentioned in Section \ref{Image-augmented Dialogue}. Their response generator \textit{Divter} has a sequence-to-sequence transformer \cite{https://doi.org/10.48550/arxiv.1706.03762} architecture, with approximately the same number of trainable parameters as \textit{GPT2-medium} and \textit{DialoGPT-medium}. The scores of PPL and BLEU-1/2 for \textit{Divter} are copied from their paper, and Distinct-1/2 is not reported. All seven models are trained and evaluated with the same dataset split in Table \ref{table:photochat_statistics}.

\begin{table}[!hbt]
  \centering
  \vspace*{15pt}
  \begin{tabular}{l|cccc}
   \hline \hline
    \textbf{Model} & \textbf{PPL$\downarrow$} & \textbf{BLEU-1/2$\uparrow$} & \textbf{Distinct-1/2$\uparrow$} \\
    \hline \hline
    \textit{Divter} & 59.63 & 0.065/0.017 & - / - \\
    \hline \hline
    \textit{GPT2-medium} & 28.30 & \textbf{0.142}/0.035 & 0.964/0.863 \\
    \hline
    \textit{DialoGPT-medium} & 27.88 & \textbf{0.142}/\textbf{0.036} & 0.969/\textbf{0.867} \\
    \hline \hline
    \textit{ViT-base+GPT2-medium} & 17.58 & 0.131/0.032 & 0.972/0.851 \\
    \hline
    \textit{ViT-base+DialoGPT-medium} & 16.86 & 0.132/0.033 & \textbf{0.976}/0.842 \\
    \hline
    \textit{ViT-large+GPT2-medium} & 17.58 & 0.132/0.033 & 0.971/0.862 \\
    \hline
    \textit{ViT-large+DialoGPT-medium} & \textbf{16.84} & 0.130/0.033 & 0.974/0.856 \\
    \hline \hline
  \end{tabular}
  \caption{Automatic evaluation results of various response generators on the test set of PhotoChat. The best score for each metric is in bold.}
  \label{table:response-generator}
\end{table}

First, in terms of PPL, all six proposed models significantly outperform \textit{Divter}. The difference in PPL between the six proposed models and \textit{Divter} is $-31.3$ at minimum and $-42.8$ at maximum, which means that the proposed models assign much higher probability to the gold responses. There are two possible interpretations. First, GPT-2 or DialoGPT is simply more effective than vanilla sequence-to-sequence transformers. Second, while \textit{Divter} does not use additional special tokens, the proposed models are trained with input and output sequences that distinguish user and bot utterances and mark the beginning and end of each sequence, as explained in Section \ref{Implementation: Response generator: Input and output}. As preliminary experiments showed that the existence and position of special tokens greatly affect the training loss, such discrepancy in PPL indicates that a careful construction of the input and output is crucial. In addition, among the six proposed models, the four multimodal response generators have much lower PPL than the two unimodal response generators by an average margin of $-10.8$, with the best absolute PPL of $16.9$. Such gap suggests that giving images as additional input to the response generator improves its language modeling capacity to predict each response in an image-augmented dialogue. Moreover, the models with DialoGPT-medium as text decoder achieve lower PPL than those with GPT2-medium, which indicates that DialoGPT is more apt in dialogue-related tasks.

In terms of BLEU-1/2, all six proposed models are also consistently better than \textit{Divter} by an approximate margin of +0.07/0.02. Combined with PPL, such gap in BLEU-1/2 supports the idea that the proposed models generate responses more similar to those of humans. Also, among the six proposed models, the four multimodal response generators only have slightly lower BLEU-1/2 compared to the two unimodal response generators by a small margin of $-0.010/0.002$. This shows that sampling tokens from text representation combined with image representation generally preserves the generation accuracy. The small margin is probably due to the fact that unimodal response generators adhere entirely to dialogue history, while multimodal generators are additionally fed with dummy images for responses that are not paired with an image, which can be small noise to the models. Not much difference in BLEU-1/2 is observed when the size of image encoder (ViT-base or ViT-large) and the type of text decoder (GPT2-medium or DialoGPT-medium) are changed. 

Furthermore, in terms of Distinct-1/2, all six proposed models achieve approximately 0.97/0.85. Such high Distinct-1/2 indicates that each response contains many unique tokens and is not a simple repetition of phrases. The four multimodal response generators achieve slightly higher Distinct-1 but lower Distinct-2 compared to the two unimodal response generators on average by a small margin of $+0.01$ and $-0.005$, respectively. There was no statistically meaningful difference in the four multimodal response generators.

\begin{figure}[!hbt]
\centering
\setlength{\belowcaptionskip}{15pt}
\includegraphics[width=1.0\linewidth]{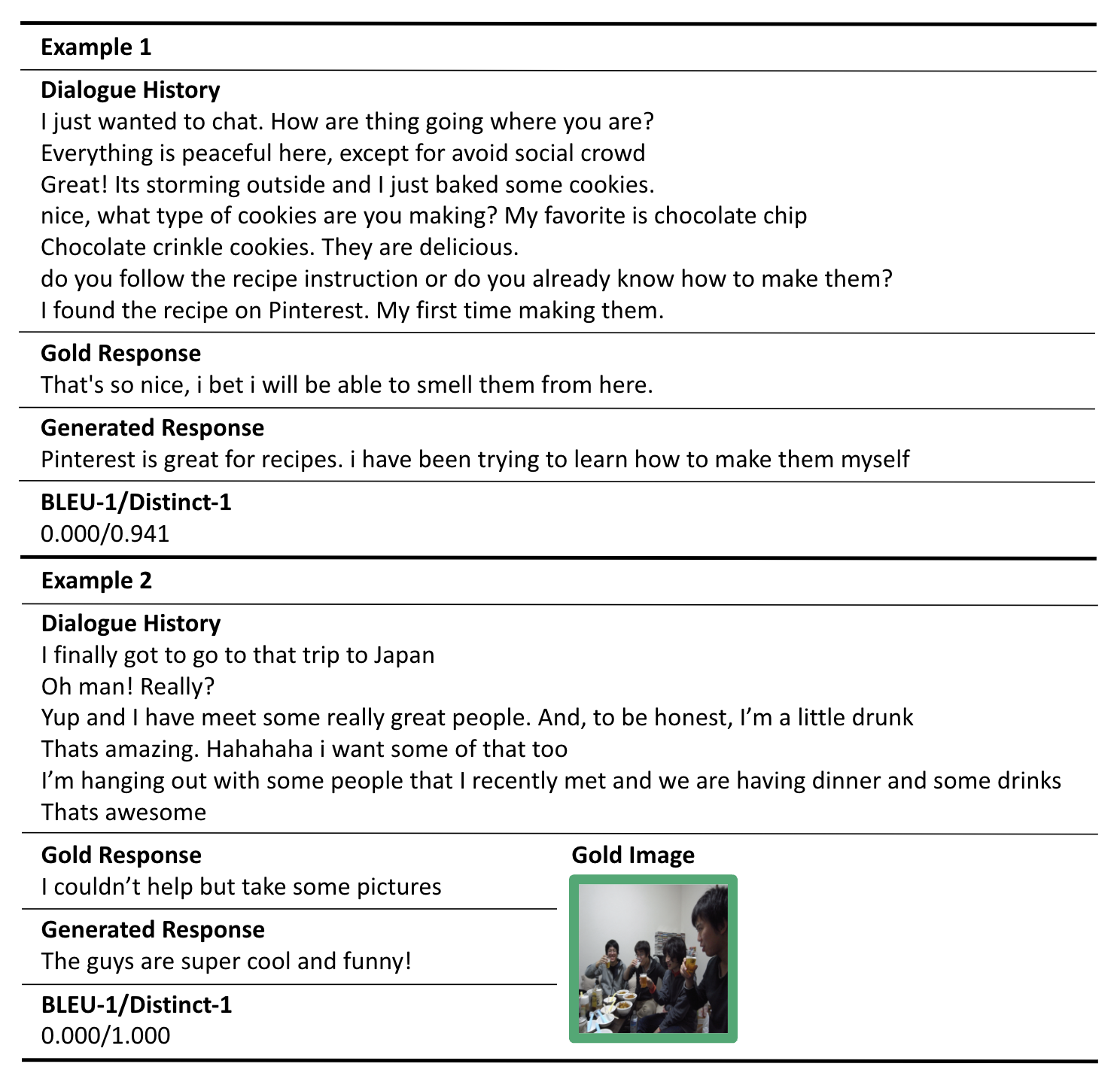}
\caption{Examples of generated responses and their automatic metric scores given each dialogue history in the test set of PhotoChat. Example 1 is a dialogue in which no image is shared. Example 2 is a dialogue in which an image is shared.}
\label{fig:sample_response_generator}
\end{figure}

Figure \ref{fig:sample_response_generator} shows the examples of responses generated by the response generator given each dialogue history in the test set of PhotoChat. The model used to generate the responses is \textit{ViT-large+DialoGPT-medium}. In Example 1, the two speakers are talking about baking cookies. Because there is no gold image, the dialogue history and a dummy image are given as model inputs. The generated response properly replies to the context by relating to the user about finding recipes on Pinterest and sharing their will to learn baking. In Example 2, one of the speakers is sharing their moment of hanging out with people they met in Japan. The dialogue history and the gold image are given as model inputs. The generated response addresses the guys in the image and is consistent with the context that the speaker is enjoying their experience in Japan. Notice that BLEU-1 is 0 in Example 1 and 2 as there are barely any overlapping tokens between the generated and the gold responses. However, each generated response is still perfectly coherent to the dialogue history and the image if applicable, and high Distinct-1 indicates that each response contains diverse tokens.

\subsection{Visualization of Model Training}

\begin{figure}[!hbt]
\centering
\setlength{\belowcaptionskip}{15pt}
\includegraphics[width=1.0\linewidth]{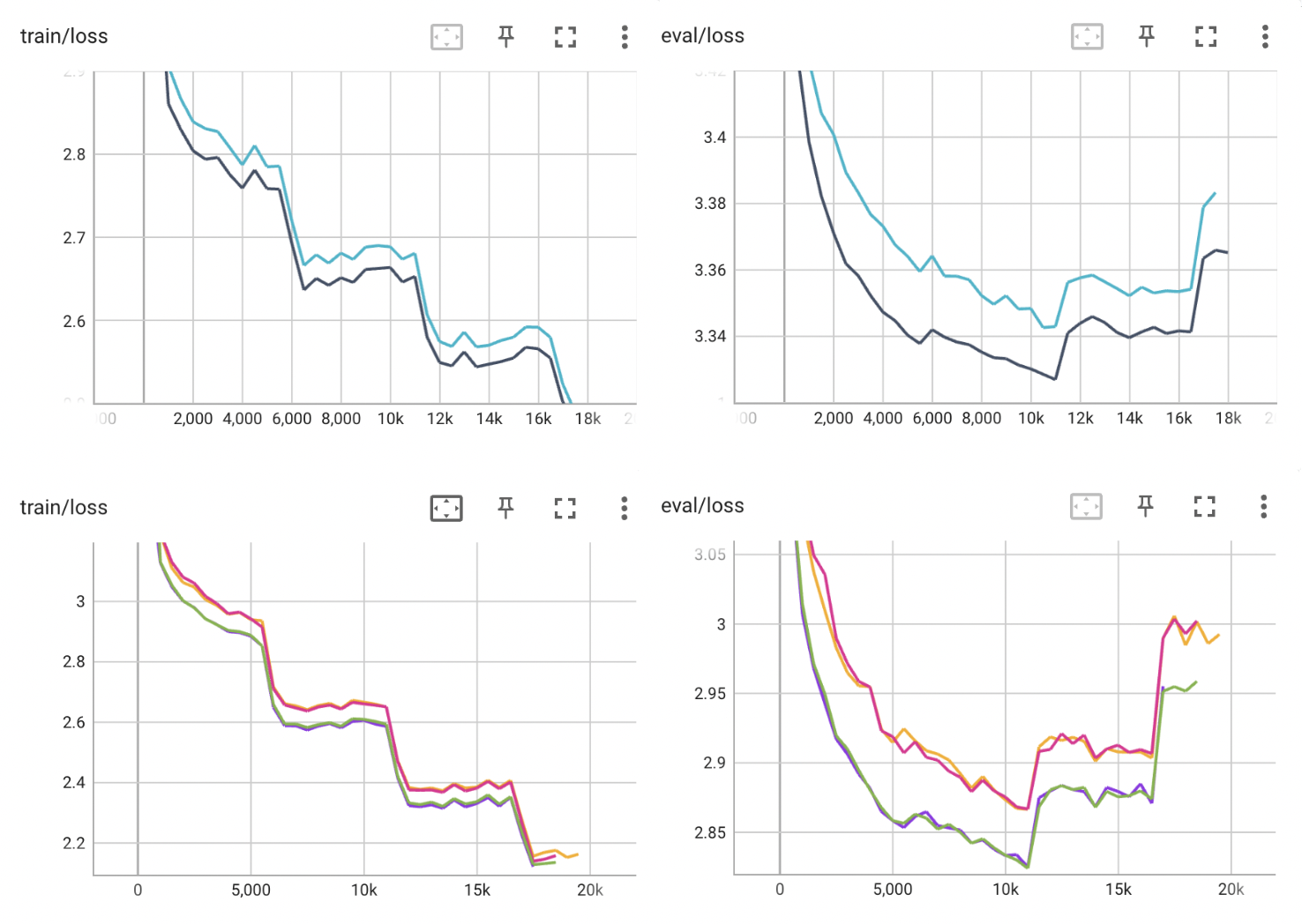}
\caption{Training and evaluation loss of response generators reported to TensorBoard. Each response generator is \textit{GPT2-medium} (blue), \textit{DialoGPT-medium} (navy), \textit{ViT-base+GPT2-medium} (yellow), \textit{ViT-base+DialoGPT-medium} (purple), \textit{ViT-large+GPT2-medium} (pink), and \textit{ViT-large+DialoGPT-medium} (green). Outliers are hidden from display.}
\label{fig:loss_response_generator}
\end{figure}

The training and evaluation losses for each response generator per training step are shown in Figure \ref{fig:loss_response_generator}. All models are trained for slightly over 3 epochs, which amounts to approximately 18,000 steps in total. At every 500 steps, the training loss is computed by averaging the loss terms from the 500 training samples. Then, the model is frozen for evaluation and is fed with all samples in the validation set, whose loss terms are averaged to compute the evaluation loss.

On the left side of Figure \ref{fig:loss_response_generator}, training loss decreases steadily for all six response generators over 3 epochs. This suggests that the cross entropy loss function does proper weight update and the models are able to converge. Specifically, the training loss drops significantly at the end of each epochs (around 6,000 steps, 11,500 steps, 17,000 steps), which locally resembles a step function.

On the right side of Figure \ref{fig:loss_response_generator}, evaluation loss decreases up to the second epoch (around 11,500 steps) but increases over the last epoch. This phenomenon is overfitting, in which evaluation loss starts increasing while training loss keeps decreasing. Thus, the best checkpoint for each response generator is chosen from the second epoch.

%On the other hand, the four multimodal response generators \textit{ViT-base+GPT2-medium}, \textit{ViT-base+DialoGPT-medium}, \textit{ViT-large+GPT2-medium}, and \textit{ViT-large+DialoGPT-medium} show steady decrease of evaluation loss over all epochs. Such difference between the evaluation losses of unimodal and multimodal response generators suggests that multimodal response generators need more epochs to converge, as the model size is larger and the task of generating response given image is more difficult than simply generating response. 

\begin{figure}[!hbt]
\centering
\setlength{\belowcaptionskip}{15pt}
\includegraphics[width=1.0\linewidth]{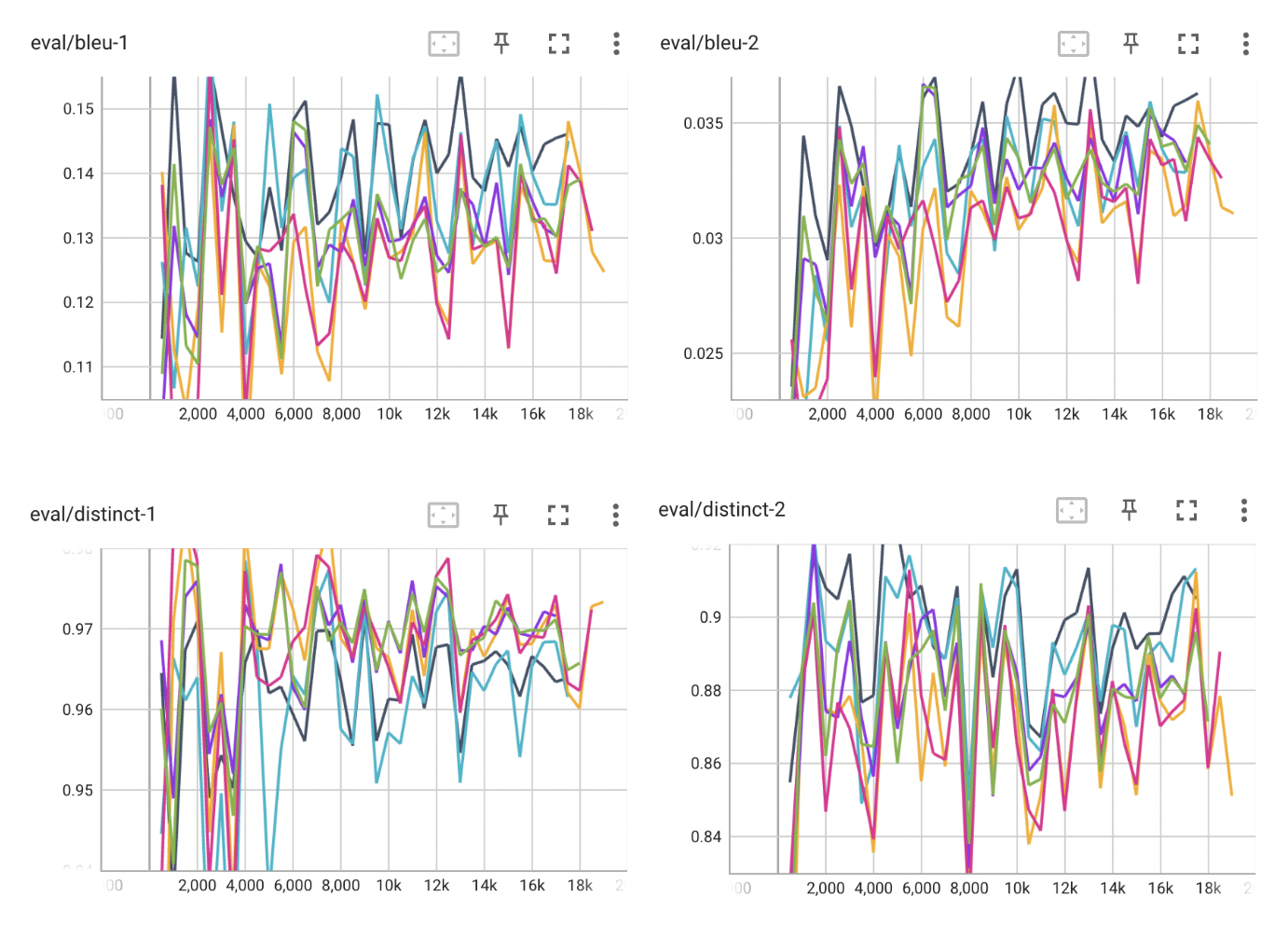}
\caption{BLEU-1/2 and Distinct-1/2 scores of response generators reported to TensorBoard. Each response generator is \textit{GPT2-medium} (blue), \textit{DialoGPT-medium} (navy), \textit{ViT-base+GPT2-medium} (yellow), \textit{ViT-base+DialoGPT-medium} (purple), \textit{ViT-large+GPT2-medium} (pink), and \textit{ViT-large+DialoGPT-medium} (green). Outliers are hidden from display.}
\label{fig:metric_response_generator}
\end{figure}

Furthermore, BLEU-1/2 and Distinct-1/2 for each response generator at each training step are shown in Figure \ref{fig:metric_response_generator}. The training and evaluation occurs every 500 steps, as in the case of computing the losses. The model under evaluation generates responses for all dialogue histories in the validation set, whose BLEU-1/2 and Distinct-1/2 are computed and averaged over the set. PPL is omitted since its curve is the same as that of the evaluation loss in Figure \ref{fig:loss_response_generator} simply in different scale. Unlike the evaluation loss that decreases steadily until certain point in Figure \ref{fig:loss_response_generator}, BLEU-1/2 and Distinct-1/2 do not change significantly from the initial training steps in Figure \ref{fig:metric_response_generator}. 

%In Figure \ref{fig:metric_response_generator}, the top figures show BLEU-1/2 and the bottom figures show Distinct-1/2. Unlike the evaluation loss and PPL that decrease at each training step in Figure \ref{fig:loss_response_generator}, BLEU-1/2 and Distinct-1/2 do not change significantly from the initial training steps. 

\subsection{Source code}

Each response generator is evaluated in \verb|class ResponseGeneratorEvaluator()| of \verb|multimodal_|\allowbreak\verb|chat/|\allowbreak\verb|learning/evaluator.py|. Here, the best model checkpoint and a callback function to compute the metrics are given as arguments to this class. This callback function is implemented in \verb|multimodal_chat/learning/callback.py|. Inside this function, for each triple of (image, dialogue history, response) in the test set, the dialogue history and a dummy image is fed to unimodal response generators, or both the dialogue history and the gold image are fed to multimodal response generators. Then, the generated response is compared in token level with the gold response to compute BLEU-1/2, and its number of unique tokens are calculated for Distinct-1/2.

When generating responses for BLEU-1/2 and Distinct-1/2, a greedy sampling strategy is used: given a probability distribution over the vocabulary, \verb|argmax| is selected as the generated token. This is to ensure that model comparison under automatic evaluation is deterministic and independent from sampling hyperparameters.

\section{Human Evaluation: Chatbot} \label{Human Evaluation: Chatbot}

In this section, the complete chatbot system consisting of image retriever and response generator is evaluated end-to-end through human evaluation. Many previous works \cite{li2019acuteeval, adiwardana2020humanlike, ji2022achieving} on chabot systems have pointed out a low correlation between automatic evaluation results and human judgement in the dialogue domain, since by nature there is no single gold response to a dialogue history. Thus, it is conventional to validate the automatic evaluation results with an interactive human evaluation procedure when testing the performance of a chatbot system.

\begin{figure}[!hbt]
\centering
\setlength{\belowcaptionskip}{15pt}
\includegraphics[width=1.0\linewidth]{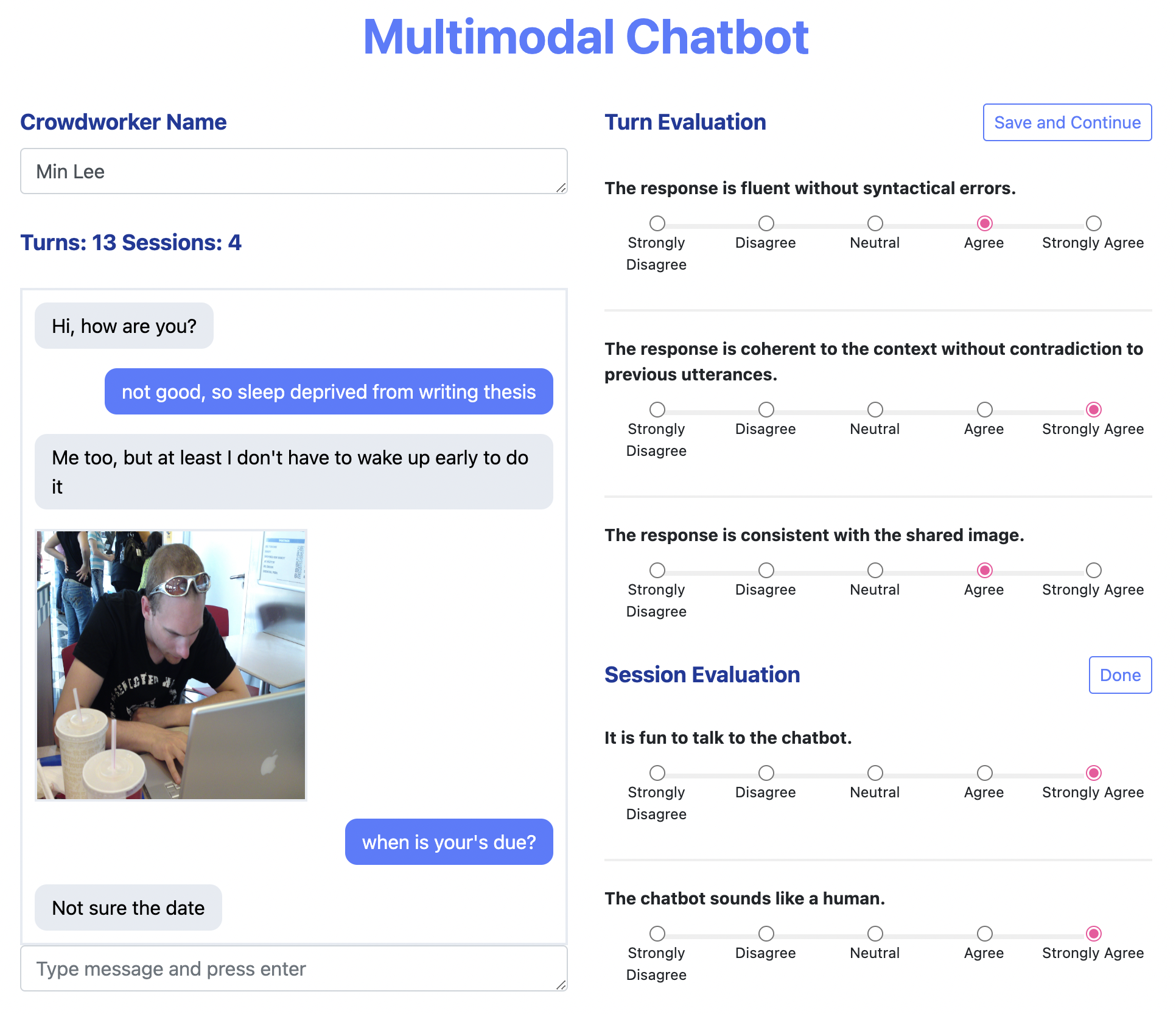}
\caption{The human evaluation interface.}
\label{fig:evaluation_interface}
\end{figure}

Human evaluation is conducted using a web interface illustrated in Figure \ref{fig:evaluation_interface}. Each crowdworker chats with the deployed chatbot and evaluates the conversation in multiple aspects. Specifically, each response from the chatbot is evaluated using turn evaluation, and once the conversation is over, the whole session is evaluated using session evaluation. 

The evaluation procedure is as follows: when each crowdworker types and sends message to the chatbot, the message is concatenated to the dialogue history, and this dialogue history is fed to the chatbot system to output a response and optionally a top 1 image. The returned response and image are concatenated to the dialogue history and displayed to the user. Then, the crowdworker indicates on a Likert scale of 1-5 how much they agree with each evaluation statement about the current turn. This turn evaluation is then repeated for about 6 turns, as the crowdworker continues the dialogue with the chatbot. Termination of dialogue is up to each crowdworker. Once the dialogue is over, the crowdworker again indicates on a Likert scale of 1-5 how much they agree with each evaluation statement about the entire session.

For evaluation, a total of 15 crowdworkers are selected from the undergraduate and graduate students of Princeton University from multiple departments. The crowdworkers are asked to chat freely with the chatbot as if they are talking to their friends. The conversation topic is up to each crowdworker as long as it does not require extreme expertise on particular fields, since the chatbot is only trained on casual dialogue dataset and does not possess accurate factual knowledge. The crowdworkers are given a thorough guideline on the evaluation procedure and are asked to play around with the web interface until they are comfortable with its functions. Each crowdworker conducts 10 dialogue sessions, and the specific model under evaluation is not revealed to the crowdworker.

\subsection{Models}

Three types of chatbot systems are evaluated for comparison of performance.

\begin{itemize}
    \item \textit{DialoGPT}: This chatbot has \textit{DialoGPT-medium} (Section \ref{Automatic Evaluation: Response Generator}) as response generator. As this chatbot does not have image retriever, it understands and responds with text only. It has approximately 355M trainable parameters. The details of response generator are described in Section \ref{Implementation: Response Generator}. 
    \item \textit{ViT+BERT+DialoGPT}: This chatbot has \textit{ViT-base+BERT-base} (Section \ref{Automatic Evaluation: Image Retriever}) as image retriever and \textit{DialoGPT-medium} (Section \ref{Automatic Evaluation: Response Generator}) as response generator. As this chatbot has unimodal generator, it understands text only and responds with image and text. It has approximately 551M trainable parameters, with 196M parameters from \textit{ViT-base+BERT-base} and 355M parameters from \textit{DialoGPT-medium}. The details of image retriever and response generator are described in Sections \ref{Implementation: Image Retriever} and \ref{Implementation: Response Generator}, respectively. 
    \item \textit{ViT+BERT+ViT+DialoGPT}: This chatbot has \textit{ViT-base+BERT-base} (Section \ref{Automatic Evaluation: Image Retriever}) as image retriever and \textit{ViT-large+DialoGPT-medium} (Section \ref{Automatic Evaluation: Response Generator}) as response generator. As this chatbot has multimodal generator, it understands and responds with both text and image. It has approximately 737M trainable parameters, with 196M parameters from \textit{ViT-base+BERT-base} and 541M parameters from \textit{ViT-large+DialoGPT-medium}. The details of image retriever and response generator are described in Sections \ref{Implementation: Image Retriever} and \ref{Implementation: Response Generator}, respectively.
\end{itemize}

\subsection{Metrics}

Given the current dialogue between each crowdworker and the chatbot, the crowdworker is presented with a set of statements that evaluate some aspect of each turn or session of the dialogue. Then, the crowdworker chooses on a Likert scale of 1-5 regarding how much they agree with each statement, where 1 is strong disagreement and 5 is strong agreement.

In turn evaluation, the three metrics are fluency, coherence, and image-groundedness:

\begin{itemize}
\item \textit{Fluency}: The response is fluent without syntactical errors.
\item \textit{Coherence}: The response is coherent to the context without contradiction to previous utterances.
\item \textit{Image-groundedness}: The response is consistent with the shared image.
\end{itemize}

In session evaluation, the two metrics are engagingness and humanness:

\begin{itemize}
\item \textit{Engagingness}: It is fun to talk to the chatbot.
\item \textit{Humanness}: The chatbot sounds like a human.
\end{itemize}

\subsection{Main Results}

This section analyzes the human evaluation performance of each chatbot system. For each of the three chatbots, a total of 50 dialogue sessions are conducted, which amounts to approximately 300 turns for evaluation.

Table \ref{table:human-eval-turn} shows the results from turn evaluation. First, the three chatbots achieve similar level of fluency and coherence around 4.1 and 3.9, respectively. As fluency and coherence are measures of basic conversational abilities, this suggests that the three models are similarly competent in carrying out dialogues and that such abilities are not significantly affected by the existence of image retriever or the type of response generator. Also, given that coherence is in the 3-point range, all three chatbots are unable to adhere perfectly to the dialogue history and may say things as if they do not even remember a couple turns ahead. Some commonly observed patterns include asking the same questions to user or misunderstanding the user's utterances as the chatbot's own. Such lack of short-term memory of chatbots have been reported even in billion-scale chatbots like Meena \cite{adiwardana2020humanlike} and LaMDA \cite{thoppilan2022lamda}. A tentative solution is to feed in as much dialogue history to response generators as possible (maximum 12 previous turns in the proposed chatbots), constrained by the maximum input length of GPT-2 and DialoGPT models. However, this problem still has not been entirely resolved and needs further research.

\begin{table}[!hbt]
  \centering
  \vspace*{15pt}
  \begin{tabular}{l|rrrrr}
   \hline \hline
    \textbf{Model} & \textbf{Fluency$\uparrow$} & \textbf{Coherence$\uparrow$} & \textbf{Image-groundedness$\uparrow$} \\
    \hline \hline
    \textit{DialoGPT} & 4.142 & 3.874 & - \\
    \hline
    \textit{ViT+BERT+DialoGPT} & 4.211 & 3.912 & 4.019 \\
    \hline
    \textit{ViT+BERT+ViT+DialoGPT} & 4.131 & 3.891 & 4.302 \\
    \hline \hline
  \end{tabular}
  \caption{Human evaluation results per turn on the proposed chatbot systems.}
  \label{table:human-eval-turn}
\end{table}

Moreover, unlike fluency and coherence, image-groundedness is higher in \textit{ViT+BERT+}\allowbreak\textit{ViT+DialoGPT} than in \textit{ViT+BERT+DialoGPT} by an approximate margin of $+0.3$. Such gap supports the initial hypothesis that the response generators that take in image as additional input will generate responses that are more consistent with the shared image, compared to the response generators that take in text only. \textit{ViT+BERT+ViT+DialoGPT} achieves an absolute score of 4.3 in image-groundedness, which signifies that it rarely says anything that contradicts the content of the shared image. However, this score may also have been overestimated, since a response that does not address anything about the image is also considered consistent. Thus, the score of 4.3 cannot be said to perfectly represent the degree of hallucination. In order to measure how much the chatbot actually hallucinates, the crowdworkers should instead ask questions to the chatbot about the shared image and mark how accurate the response is, which is not the scope of this human evaluation. Image-groundedness is not measured in \textit{DialoGPT} since it does not have any image retriever.

Table \ref{table:human-eval-session} shows the results from session evaluation. Among the three chatbots, \textit{ViT+}\allowbreak\textit{BERT+}\allowbreak\textit{ViT+DialoGPT} achieves the highest engagingness with a score of 4.3. Its score is higher than that of \textit{ViT+BERT+DialoGPT} approximately by $+0.1$, which may suggest that a response generator that understands both image and text can make the users slightly more interested in the conversation, since the user will get less distracted by inconsistent responses. Also, the engagingness of \textit{ViT+BERT+ViT+DialoGPT} is much higher than that of \textit{DialoGPT} by $+0.6$, which shows that using an image retriever in a chatbot to share images can substantially improve the user experience of talking to a chatbot.

\begin{table}[!hbt]
  \centering
  \vspace*{15pt}
  \begin{tabular}{l|rrrrr}
   \hline \hline
    \textbf{Model} & \textbf{Engagingness$\uparrow$} & \textbf{Humanness$\uparrow$} \\
    \hline \hline
    \textit{DialoGPT} & 3.720 & 3.169 \\
    \hline
    \textit{ViT+BERT+DialoGPT} & 4.184 & 3.057 \\
    \hline
    \textit{ViT+BERT+ViT+DialoGPT} & 4.299 & 3.118 \\
    \hline \hline
  \end{tabular}
  \caption{Human evaluation results per session on the proposed chatbot systems.}
  \label{table:human-eval-session}
\end{table}

Furthermore, all three chatbots achieve roughly 3.1 of humanness, with the highest score from \textit{DialoGPT}. However, the difference in performance among the three chatbots is not as significant as that in engagingness. This suggests that the existence of image retriever or the type of response generator does not largely impact how much a chatbot resembles a human. Given that the scores are around 3.1, the proposed chatbot systems are quite far from sounding perfectly like humans, sometimes due to qualities that are not explicitly definable with a metric. Even if a response from chatbot is completely fluent and coherent, it might still sound awkward, making it come across as a bot. Also, a qualitative look over the collected dialogues reveals that while the chatbot is generally able to understand common slangs or acronyms used by the crowdworkers like ``hbu" or ``yayyy," it rarely generates such tokens in the responses, since their probabilities are presumably very low in the pretrained corpus. Such lack of informality may also harm the humanness of the chatbot.

\begin{figure}[!hbt]
\centering
%\vspace*{15pt}
\setlength{\belowcaptionskip}{15pt}
\includegraphics[width=1.0\linewidth]{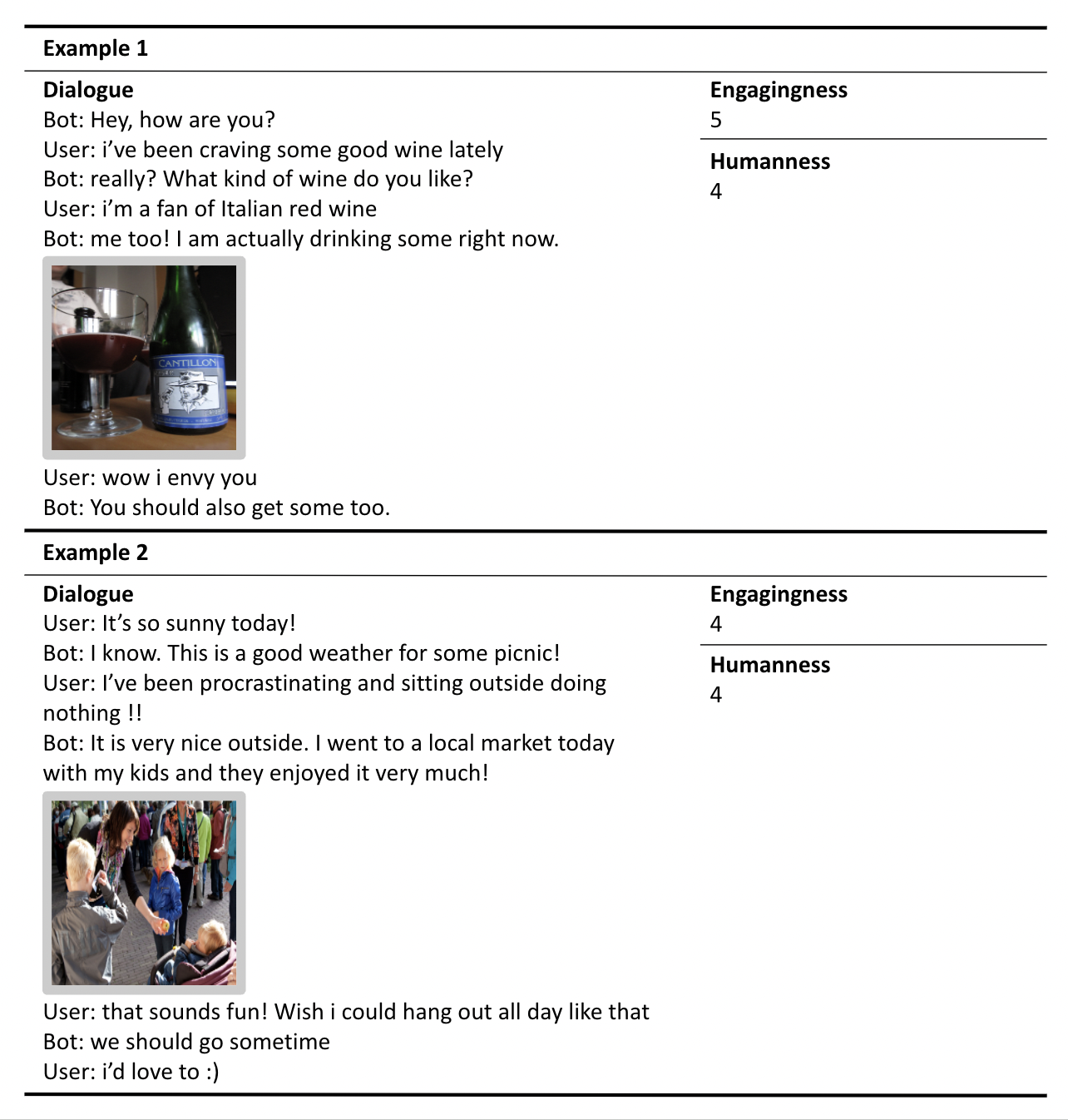}
\caption{Sample dialogues conducted by crowdworkers.}
\label{fig:dialogue_sample}
\end{figure}

Examples of cherry-picked dialogues carried out by crowdworkers are shown in Figure \ref{fig:dialogue_sample}. The chatbot system that retrieved the images and generated the responses is \textit{ViT+BERT+ViT+DialoGPT}. In Example 1, the user mentions that they like red wine, and the bot responds with a photo of wine and says that they are drinking it at the moment. Here, the engagingness of this session is rated as 5, a perfect score since the bot makes the user engaged by empathizing and asking questions about the user's interest directly. The humanness of the session is 4, presumably because it seems a little abrupt that the bot happens to be drinking wine at the moment. In Example 2, the bot and the user talk about a sunny weather, and the bot shares a picture of a local market they went to. The engagingness and humanness scores are both 4, as the conversation goes on smoothly but is not particularly mindblowing. Aside from the evaluation scores, Example 2 also demonstrates two main limitation that chatbots have. First, the bot hallucinates about the sunny weather and obliviously agrees with the user, even though the bot does not have any notion of current time or location. Also, the bot offers the user to go to the market together, but the bot and the user cannot meet in reality.

\subsection{Source code}

The human evaluation interface in Figure \ref{fig:evaluation_interface} is implemented in \verb|multimodal_chat/demo| using HTML/JavaScript for frontend and Flask for backend. In this directory, \verb|multimodal_chat/|\allowbreak\verb|demo/run_demo.py| is the main script that loads pages from \verb|multimodal_chat/demo/|\allowbreak\verb|templates| and interactions and styles from \verb|multimodal_chat/|\allowbreak\verb|demo/static|. Each dialogue and evaluation results are saved as a json file.

When generating responses for human evaluation, a stochastic sampling strategy is used. Specifically, nucleus sampling \cite{holtzman2020curious} is used with top-$p$ of 0.1, which means that only the tokens with initial probability of 0.1 are considered, and a token is sampled from the rescaled distribution of these probabilities. This is to ensure that the chatbot responses are diverse while controlled to certain extent.

\chapter{Conclusion}
\section{Summary}

The objective of this work is to build a chatbot that can converse with users about daily topics while sending relevant photos when appropriate. The key novelty of this work is that the components of the chatbot are multimodal: the image retriever and the response generator understand both image and text. The effectiveness of each model is independently validated using various automatic evaluation metrics such as Recall@1/5/10, MRR, PPL, BLEU-1/2, and Distinct-1/2. Furthermore, the full proposed chatbot with multimodal image retriever and response generator is demonstrated to achieve higher image-groundedness and engagingness, along with competitive fluency, coherence, and humanness, when compared to other chatbot variants in human evaluation. This work is a meaningful step towards developing an interactive AI chatbot that leverages both image and text modalities to engage in human-like conversations. 

\section{Limitations}

One limitation of this work is that the proposed image retriever selects images from a finite set. Although the image database in PhotoChat \cite{zang-etal-2021-photochat} contains more than 10,000 images from various domains, it still may not cover long-tail images such as a tiger walking to Fruity Yogurt. One possible solution is to add an image translator that generates images online, as implemented in one of the previous works \cite{sun-etal-2022-multimodal}. However, a remaining issue that has not been addressed by both the previous work \cite{sun-etal-2022-multimodal} and the proposed chatbot is that images are not constrained by any persona information of the chatbot. For example, even when the persona of the chatbot is fixed as a teenage girl, the chatbot may send an image of a White man and still generate something like ``Look what I'm doing right now." Future works could explore mechanisms for controlling multiple aspects of retrieved or generated images. One possible approach would be DreamBooth \cite{https://doi.org/10.48550/arxiv.2208.12242}, a recent work on text-to-image diffusion models for generating images with a fixed target object. This method would enable a chatbot with an explicit persona, such as a visual avatar, to generate images of the avatar in various scenarios.

This work is additionally constrained by the lack of out-of-distribution (OOD) detection mechanism for image retrieval. The proposed chatbot system relies on an empirically set threshold that decides whether the top 1 image should be retrieved. Even though the automatic evaluation performance of each model is independent from this threshold, the human evaluation results might be sensitive to it, since it affects how often images are shared to users. However, this decision boundary is vague by nature, since there is no correct answer to when a photo should be shared in an open dialogue. One previous work \cite{zang-etal-2021-photochat} attempts to solve this issue using a photo-sharing intent classifier, but this model is reported to achieve an F1 score of only 58.9 in binary classification. A more complicated OOD detection algorithm such as MC dropout \cite{gal2016dropout} can be studied in the dialogue domain and be incorporated into this work.

Moreover, the proposed multimodal response generator is limited to processing only one image at a time. This limitation stems from the design of the vision-text dual encoder, which takes in only a single image input. However, if multiple images are shared in a dialogue, the response generator should ideally understand all of them and generate responses accordingly. A possible future work might be implementing response generators that can accept multiple image inputs while being able to semantically distinguishing them.

A further limitation of this work is that the sharing of images occurs asymmetrically: while the chatbot is able to send both messages and photos, it can receive only messages from users. The current version of the web interface does not allow users to send photos themselves, as it is designed solely for evaluating the chatbot performance in retrieving correct images and generating proper responses conditioned on those images. Additionally, PhotoChat only contains dialogues in which one speaker shares images, whereas multiple speakers share images in real-life conversations. Therefore, a natural direction for further research would be building a chatbot system that understands messages and photos from an undefined number of speakers.

\appendix
\chapter{Code}
\dirtree{%
 .1 multimodal\_chat.
 .2 dataset.
 .3 collator.py.
 .3 processor.py.
 .2 demo.
 .3 static.
 .4 custom.css.
 .4 script.js.
 .3 templates.
 .4 evaluation.html.
 .4 main.html.
 .3 config.py.
 .3 run\_demo.py.
 .2 learning.
 .3 callback.py.
 .3 evaluator.py.
 .3 trainer.py.
 .2 model.
 .3 image\_retriever.py.
 .3 response\_generator.py.
 .2 sh.
 .3 eval\_image\_retriever.sh.
 .3 eval\_response\_generator.sh.
 .3 train\_image\_retriever.sh.
 .3 train\_response\_generator.sh.
 .2 util.
 .3 args.py.
 .3 image.py.
 .3 io.py.
 .3 io.py.
 .3 metric.py.
 .3 resource.py.
 .3 text.py.
 .3 time.py.
 .2 README.md.
 .2 requirements.txt.
 .2 run\_image\_retriever.py.
 .2 run\_response\_generator.py.
 .2 setup.sh.
}

\chapter{Model Architectures}
\label{Appendix: Model architecture}

\begin{table}[hbt]
  \centering
  \begin{tabular}{l|rrrr}
   \hline \hline
    \textbf{Model} & \textbf{\# parameters} & \textbf{\# layers} & \textbf{\# hidden size} & \textbf{\# heads} \\
    \hline \hline
    ViT-base & 86M & 12 & 768 & 12\\
    \hline
    ViT-large & 307M &  24 & 1024 & 16\\
    \hline
    BERT-base & 110M & 12 & 768 & 12\\
    \hline
    BERT-large & 336M &  24 & 1024 & 16\\
    \hline \hline
  \end{tabular}
  \caption{Details of each model architecture for image retrievers.}
  \label{table:image-retriever-architecture}
\end{table}

\begin{table}[hbt]
  \centering
  \begin{tabular}{l|rrrr}
   \hline \hline
    \textbf{Model} & \textbf{\# parameters} & \textbf{\# layers} & \textbf{\# hidden size} & \textbf{\# heads} \\
    \hline \hline
    ViT-base & 86M & 12 & 768 & 12\\
    \hline
    ViT-large & 307M &  24 & 1024 & 16\\
    \hline
    GPT2-medium & 345M &  24 & 1024 & 16\\
    \hline
    DialoGPT-medium & 345M &  24 & 1024 & 16\\
    \hline \hline
  \end{tabular}
  \caption{Details of each model architecture for response generators.}
  \label{table:response-generator-architecture}
\end{table}

\bibliographystyle{abbrv}
\bibliography{refs} \label{bib}

\end{document}